\def\arxivversion{1}
\documentclass{article}
\providecommand{\arxivversion}{0}
\newif\ifblindvault
\blindvaultfalse
\usepackage[utf8]{inputenc}
\usepackage[T1]{fontenc}
\usepackage[expansion=false,protrusion=true]{microtype}
\usepackage{graphicx}
\usepackage{booktabs}
\usepackage{hyperref}
\ifnum\arxivversion=1
\usepackage[preprint]{icml2026}
\hypersetup{pdfsubject={}}
\else
\usepackage{icml2026}
\fi
\usepackage{xurl}
\usepackage{amsmath,amssymb,amsthm}
\usepackage{xcolor}
\usepackage[capitalise,noabbrev]{cleveref}
\newcommand{\ajHThreeB}{12}
\newcommand{\ajHThreeP}{.0005}
\newcommand{\astraAuditN}{20}
\newcommand{\astraAuditOver}{8}
\newcommand{\astraHumanLabel}{26}
\newcommand{\astraHumanRetire}{29}

\newcommand{\aucSTwo}{0.963}
\newcommand{\aucSThree}{0.957}

\newcommand{\aucSEight}{0.877}

\newcommand{\aucSSeven}{0.823}
\newcommand{\bandAgreeMaj}{32}
\newcommand{\bandAgreeOne}{26}
\newcommand{\bandAgreeThree}{28}
\newcommand{\bandAgreeTwo}{35}
\newcommand{\cpKMFour}{2}
\newcommand{\cpNeedN}{40}
\newcommand{\cpUpMFour}{18.9}
\newcommand{\csCos}{.29}
\newcommand{\ddetMFour}{94.4}
\newcommand{\ddetSTwoSeventy}{77.8}

\newcommand{\detSTwo}{93.1}
\newcommand{\detSThree}{77.8}

\newcommand{\detSThreeX}{51.4}
\newcommand{\detSFour}{80.6}
\newcommand{\detSEight}{45.8}
\newcommand{\detSEightRaw}{66.7}

\newcommand{\detSFiveR}{62.5}
\newcommand{\detSSeven}{6.9}
\newcommand{\detSSix}{33.3}
\newcommand{\detSSixC}{25.0}
\newcommand{\detSSixDel}{20.8}
\newcommand{\detSSixX}{56.9}

\newcommand{\detSThreeTOne}{81.4}
\newcommand{\detSThreeTTwo}{61.5}
\newcommand{\detSTwoTOne}{96.6}
\newcommand{\detSTwoTTwo}{76.9}

\newcommand{\detTauSixtyFive}{79.2}

\newcommand{\devBandN}{30}

\newcommand{\dpreMFour}{6.5}
\newcommand{\dpreSTwoSeventy}{9.7}
\newcommand{\ebA}{72}
\newcommand{\ebB}{68}
\newcommand{\ebCComp}{30}
\newcommand{\ebCSup}{30}
\newcommand{\ebCards}{210}
\newcommand{\ebItems}{200}
\newcommand{\ebRep}{10}
\newcommand{\extAOne}{23}
\newcommand{\extAThree}{54}
\newcommand{\extATwo}{44}
\newcommand{\extMaj}{48}
\newcommand{\extMedianS}{6}
\newcommand{\extSpanMin}{11.5}
\newcommand{\fcKSFour}{2}
\newcommand{\fcKSThree}{3}
\newcommand{\fcKSTwo}{4}
\newcommand{\fcN}{18}

\newcommand{\fiSFiveR}{4.2}

\newcommand{\fiSSixC}{0.6}

\newcommand{\hNCAstra}{19}
\newcommand{\hNCN}{25}
\newcommand{\hNCSEight}{24.0}
\newcommand{\hNCSEightK}{6}
\newcommand{\hNCSFive}{32.0}
\newcommand{\hNCSThree}{48.0}
\newcommand{\hNCSTwo}{80.0}
\newcommand{\hThreeB}{24}
\newcommand{\hThreeC}{1}
\newcommand{\hThreeLoss}{15.3}
\newcommand{\hThreeP}{1.5\times 10^{-6}}
\newcommand{\hTwoHuman}{9}
\newcommand{\hTwoN}{11}
\newcommand{\hTwoSFour}{63.6}
\newcommand{\hTwoSTwo}{90.9}
\newcommand{\hTwoUnan}{2}
\newcommand{\jointAuthCodex}{58}
\newcommand{\jointExtAuth}{52}
\newcommand{\jointExtCodex}{47}
\newcommand{\kappaOneThree}{0.22}
\newcommand{\kappaOneTwo}{0.28}
\newcommand{\kappaTwoThree}{0.60}
\newcommand{\labAuthCodex}{67}
\newcommand{\labExtAuth}{64}
\newcommand{\labExtCodex}{65}
\newcommand{\mNCN}{72}
\newcommand{\mNCSFive}{37.5}
\newcommand{\mNCSThree}{4.2}
\newcommand{\mNCSTwo}{25.0}

\newcommand{\mxCA}{0}
\newcommand{\mxCB}{2}
\newcommand{\mxCP}{.50}
\newcommand{\mxCalls}{2,340}
\newcommand{\mxDetA}{2}
\newcommand{\mxDetB}{1}
\newcommand{\mxDetP}{1.00}
\newcommand{\mxDiff}{+1.0}
\newcommand{\mxDiffHi}{+5.3}
\newcommand{\mxDiffLo}{$-$3.1}

\newcommand{\mxMcA}{2}
\newcommand{\mxMcB}{1}
\newcommand{\mxMcP}{1.00}

\newcommand{\mxRepN}{169}
\newcommand{\mxRepThree}{164}
\newcommand{\mxRepThreePreK}{16}
\newcommand{\mxRepYN}{168}
\newcommand{\mxRepYNPreK}{18}
\newcommand{\mxSTwoA}{20}
\newcommand{\mxSTwoB}{1}

\newcommand{\mxSTwoPreK}{37}
\newcommand{\mxShare}{5}
\newcommand{\mxStabN}{500}
\newcommand{\mxStabSTwo}{493}
\newcommand{\mxStabThree}{497}

\newcommand{\mxTau}{.52}
\newcommand{\mxTauA}{1}
\newcommand{\mxTauB}{2}
\newcommand{\mxTauDetK}{56}
\newcommand{\mxTauP}{1.00}
\newcommand{\mxTauPreK}{16}

\newcommand{\mxThreeDetK}{56}
\newcommand{\mxThreePre}{17.5}
\newcommand{\mxThreePreHi}{26.3}
\newcommand{\mxThreePreK}{17}
\newcommand{\mxThreePreLo}{11.2}
\newcommand{\mxUSD}{0.08}
\newcommand{\mxWsN}{141}

\newcommand{\mxYNDetK}{57}
\newcommand{\mxYNPre}{18.6}
\newcommand{\mxYNPreHi}{27.4}
\newcommand{\mxYNPreK}{18}
\newcommand{\mxYNPreLo}{12.1}

\newcommand{\mxYNTwoPreK}{15}
\newcommand{\nBand}{39}
\newcommand{\nBatchOne}{41}
\newcommand{\nChangeAgree}{228}
\newcommand{\nChangeTest}{179}
\newcommand{\nCrossTest}{179}

\newcommand{\nDevAstra}{16}
\newcommand{\nDevAstraBand}{15}
\newcommand{\nDevAstraNC}{8}
\newcommand{\nDevAstraSup}{8}
\newcommand{\nDevCand}{26}
\newcommand{\nDevNC}{31}
\newcommand{\nDevSup}{18}
\newcommand{\nHumanTest}{72}
\newcommand{\nNeighbourComp}{313}

\newcommand{\nReadj}{68}
\newcommand{\nReadjTest}{31}
\newcommand{\nTOne}{89}
\newcommand{\nTTwo}{90}
\newcommand{\nTest}{500}
\newcommand{\nTestComp}{331}
\newcommand{\nTestNC}{97}
\newcommand{\nTestSup}{72}
\newcommand{\nVault}{176}
\newcommand{\nVaultChange}{63}
\newcommand{\nVaultCross}{63}
\newcommand{\nVaultHuman}{37}
\newcommand{\nVaultNC}{41}
\newcommand{\nVaultSup}{20}
\newcommand{\nVaultWithin}{50}
\newcommand{\nWithinComp}{141}
\newcommand{\nWithinTest}{142}
\newcommand{\nameN}{3.6}
\newcommand{\nameX}{10.2}
\newcommand{\ncFiveOnlyTwo}{19}
\newcommand{\ncSTwoOnlyFive}{22}

\newcommand{\ncTTwo}{66}

\newcommand{\noCallSSix}{40}
\newcommand{\pdDivDetS}{80.6}
\newcommand{\pdDivDetX}{59.7}
\newcommand{\pdDivDisX}{11.4}
\newcommand{\pdPriorSupS}{.14}
\newcommand{\pdPriorSupX}{.11}

\newcommand{\pdRawDisX}{10.2}

\newcommand{\preSTwo}{39.2}
\newcommand{\preSThree}{15.5}

\newcommand{\preSFour}{14.6}
\newcommand{\preSEight}{12.4}
\newcommand{\preSEightHi}{20.4}
\newcommand{\preSEightLo}{7.2}
\newcommand{\preSEightRaw}{35.1}
\newcommand{\preSFive}{36.1}
\newcommand{\preSFiveHi}{46.0}
\newcommand{\preSFiveLo}{27.2}
\newcommand{\preSFiveR}{14.4}

\newcommand{\preSSix}{8.2}
\newcommand{\preSSixC}{6.2}
\newcommand{\preSSixDel}{6.2}

\newcommand{\preSSixX}{14.4}

\newcommand{\preSThreeCh}{16.9}

\newcommand{\preSThreeHi}{24.0}
\newcommand{\preSThreeLo}{9.6}

\newcommand{\preTauSixtyFive}{17.5}
\newcommand{\pwOneThree}{29}
\newcommand{\pwOneTwo}{19}
\newcommand{\pwTwoThree}{42}
\newcommand{\rAEditSThree}{3}
\newcommand{\rAEditSTwo}{27}
\newcommand{\rAOrigSThree}{39}
\newcommand{\rAOrigSTwo}{47}
\newcommand{\rAPlanN}{45}

\newcommand{\rAucDiff}{+.014}
\newcommand{\rAucHi}{+.048}
\newcommand{\rAucLo}{$-$.020}
\newcommand{\rAucP}{.41}
\newcommand{\rAucSThree}{.880}
\newcommand{\rAucSTwo}{.865}
\newcommand{\rBEditSThree}{46}
\newcommand{\rBEditSTwo}{49}
\newcommand{\rBOrigSThree}{8}
\newcommand{\rBOrigSTwo}{21}
\newcommand{\rDetSThree}{56}
\newcommand{\rDetTau}{55}

\newcommand{\rFlipOnlyThree}{34}
\newcommand{\rFlipOnlyTwo}{8}
\newcommand{\rFlipP}{.0001}
\newcommand{\rFlipSThree}{74}
\newcommand{\rFlipSTwo}{48}
\newcommand{\rLR}{15.1}
\newcommand{\rLRP}{.0001}
\newcommand{\rLRRevP}{.27}
\newcommand{\rMcA}{10}
\newcommand{\rMcB}{11}
\newcommand{\rMcP}{1.00}
\newcommand{\rMcSensP}{.66}

\newcommand{\rPairs}{100}
\newcommand{\rPairsDir}{50}

\newcommand{\rPreTau}{15}
\newcommand{\rSkipA}{4}
\newcommand{\rSkipB}{8}
\newcommand{\rStableSThree}{100}
\newcommand{\rStableSTwo}{95}
\newcommand{\rTau}{.66}
\newcommand{\rTauLo}{.64}
\newcommand{\restAuditCtrl}{3}
\newcommand{\restAuditN}{20}
\newcommand{\restAuditOver}{5}
\newcommand{\restAuditRetire}{1}
\newcommand{\restAuditToComp}{1}
\newcommand{\retExtAuth}{67}
\newcommand{\retExtCodex}{68}
\newcommand{\retExtCodexVault}{37}
\newcommand{\rtDays}{20}
\newcommand{\rtFlipK}{3}

\newcommand{\rtFlipPct}{6.0}
\newcommand{\rtHThreeMargin}{4}
\newcommand{\rtMaxAbs}{.13}
\newcommand{\rtN}{50}
\newcommand{\rvAucHi}{+.085}
\newcommand{\rvAucLo}{$-$.045}
\newcommand{\rvLRP}{.096}
\newcommand{\rvMcA}{5}
\newcommand{\rvMcB}{4}
\newcommand{\sFourCov}{179}
\newcommand{\sFourNC}{89}
\newcommand{\sbBfLTwoOnlyD}{10}
\newcommand{\sbBfLZeroLTwoDP}{.039}
\newcommand{\sbBfLZeroOnlyD}{2}
\newcommand{\sbBoth}{12}
\newcommand{\sbCertNC}{19.2}
\newcommand{\sbCertNCK}{16}
\newcommand{\sbCertNCN}{122}
\newcommand{\sbComp}{20}
\newcommand{\sbDoseDOne}{100.0}
\newcommand{\sbDoseDZero}{45.8}
\newcommand{\sbDoseDraws}{5}
\newcommand{\sbDoseN}{118}
\newcommand{\sbDosePHalf}{61.0}
\newcommand{\sbDosePOne}{85.4}
\newcommand{\sbDosePQuarter}{48.3}
\newcommand{\sbDosePZero}{34.1}
\newcommand{\sbFirstHeadK}{5}
\newcommand{\sbFirstMajK}{4}
\newcommand{\sbFirstN}{41}
\newcommand{\sbFirstNC}{15}
\newcommand{\sbFisherHolm}{.45}
\newcommand{\sbFisherP}{.22}
\newcommand{\sbFiveD}{60.4}

\newcommand{\sbFiveOverlap}{8}
\newcommand{\sbFiveP}{41.5}
\newcommand{\sbFivePK}{17}
\newcommand{\sbGFourK}{20}
\newcommand{\sbGFourKBf}{19}
\newcommand{\sbGFourN}{20}
\newcommand{\sbGH}{24}
\newcommand{\sbGOneAcc}{.895}
\newcommand{\sbGOneBase}{.662}
\newcommand{\sbGTwo}{.759}
\newcommand{\sbHeadP}{34.1}
\newcommand{\sbHeadPHi}{49.5}
\newcommand{\sbHeadPK}{14}
\newcommand{\sbHeadPLo}{21.6}
\newcommand{\sbLTwoOnlyD}{7}
\newcommand{\sbLTwoOnlyP}{7}
\newcommand{\sbLZeroLTwoDP}{.18}
\newcommand{\sbLZeroLTwoPP}{.34}
\newcommand{\sbLZeroOnlyD}{2}
\newcommand{\sbLZeroOnlyP}{3}
\newcommand{\sbMajP}{31.7}
\newcommand{\sbMajPK}{13}
\newcommand{\sbMajQNK}{17}
\newcommand{\sbN}{109}
\newcommand{\sbNC}{41}
\newcommand{\sbNTest}{72}
\newcommand{\sbNVault}{37}
\newcommand{\sbPool}{532}
\newcommand{\sbPoolLoneK}{4}
\newcommand{\sbPoolLoneN}{92}
\newcommand{\sbSG}{41}
\newcommand{\sbSH}{41}
\newcommand{\sbShareA}{6}
\newcommand{\sbShareB}{8}
\newcommand{\sbShareNOne}{13}
\newcommand{\sbShareNTwo}{28}
\newcommand{\sbStaleDisc}{21}
\newcommand{\sbStaleDiscB}{0}
\newcommand{\sbStaleK}{38}
\newcommand{\sbStaleP}{4.8\times 10^{-7}}
\newcommand{\sbSup}{48}
\newcommand{\sbTau}{.158}
\newcommand{\sensBOneLoss}{15.5}
\newcommand{\sensBOneNCN}{102}
\newcommand{\sensBOneSThreePre}{16.7}
\newcommand{\sensBOneSTwoPre}{39.2}
\newcommand{\sensNoLoss}{12.7}
\newcommand{\sensNoNCN}{100}
\newcommand{\sensNoSThreePre}{17.0}
\newcommand{\sensNoSTwoPre}{41.0}
\newcommand{\shBoth}{7}
\newcommand{\shGH}{11}
\newcommand{\shOver}{9}
\newcommand{\shSG}{7}
\newcommand{\shSH}{10}
\newcommand{\stUnused}{105}
\newcommand{\stalePairsEstablished}{72}

\newcommand{\supTOne}{59}
\newcommand{\supTTwo}{13}
\newcommand{\swSSevenD}{31.9}

\newcommand{\uidDev}{49}
\newcommand{\uidTest}{179}
\newcommand{\uidVault}{63}
\newcommand{\updSSixN}{20}
\newcommand{\updSSixSame}{16}

\newcommand{\vAgreeThree}{30}

\newcommand{\vBandAgree}{17}
\newcommand{\vBandN}{27}
\newcommand{\vDiffNC}{4}
\newcommand{\vDiffNCEach}{2}
\newcommand{\vDiffSup}{2}
\newcommand{\vDiffSupEach}{1}
\newcommand{\vHumNCN}{41}
\newcommand{\vHumPreMFourK}{11}

\newcommand{\vMajNCN}{37}
\newcommand{\vMajPreMFour}{18.9}
\newcommand{\vMajPreMFourK}{7}

\newcommand{\vNCHumK}{16}
\newcommand{\vNCMajK}{25}
\newcommand{\vSThreeGapDays}{17}

\newcommand{\vaultBand}{22.2}
\newcommand{\vdetMFive}{95.0}
\newcommand{\vdetMFour}{95.0}
\newcommand{\vdetSEight}{60.0}
\newcommand{\vdetSFive}{70.0}
\newcommand{\vdetSSeven}{10.0}
\newcommand{\vdetSSix}{30.0}
\newcommand{\vdetSSixC}{25.0}

\newcommand{\vdetSThreeV}{95.0}
\newcommand{\vdetSThreeXV}{55.0}

\newcommand{\vhThreeA}{9}
\newcommand{\vhThreeB}{1}
\newcommand{\vhThreeLoss}{5.0}
\newcommand{\vhThreeOneLessP}{.039}
\newcommand{\vhThreeP}{.021}
\newcommand{\vhThreeStep}{5}
\newcommand{\vhThreeTwoLessP}{.070}
\newcommand{\vmcA}{7}
\newcommand{\vmcB}{0}
\newcommand{\vmcP}{.016}
\newcommand{\vpreMFive}{26.8}
\newcommand{\vpreMFour}{26.8}
\newcommand{\vpreSEight}{19.5}
\newcommand{\vpreSEightK}{8}
\newcommand{\vpreSFive}{51.2}
\newcommand{\vpreSFiveR}{31.7}
\newcommand{\vpreSSeven}{2.4}
\newcommand{\vpreSSix}{12.2}
\newcommand{\vpreSSixC}{7.3}

\newcommand{\vpreSThreeNV}{24.4}
\newcommand{\vpreSThreeV}{24.4}
\newcommand{\vpreSThreeVHi}{39.3}
\newcommand{\vpreSThreeVLo}{13.8}
\newcommand{\vpreSThreeXV}{14.6}
\newcommand{\vpreSTwoV}{43.9}
\newcommand{\vwsN}{50}
\newcommand{\vwsSSixCK}{0}
\newcommand{\vwsSSixK}{3}

\newcommand{\wsCos}{.39}

\newcommand{\wsSOne}{21.3}

\newcommand{\wsSEightRaw}{10.6}
\newcommand{\wsSFive}{11.3}
\newcommand{\wsSFiveK}{16}
\newcommand{\wsSFiveR}{1.4}

\newcommand{\wsSSix}{6.4}

\newcommand{\wsSSixCK}{1}
\newcommand{\wsSSixK}{9}
\newcommand{\wsSSixNoop}{8}
\newcommand{\wsSSixX}{8.5}
\newcommand{\wsSSixXC}{5.7}
\newcommand{\wsSSixXCK}{8}
\newcommand{\wsSSixXK}{12}
\newcommand{\wsSSixXNoop}{4}

\newtheorem{proposition}{Proposition}
\definecolor{supc}{HTML}{12807A}
\definecolor{ncc}{HTML}{C4553B}
\definecolor{compc}{HTML}{A2770E}
\newcommand{\SUP}{\textup{\textsc{\color{supc}superseded}}}
\newcommand{\NC}{\textup{\textsc{\color{ncc}needs-confirmation}}}
\newcommand{\COMP}{\textup{\textsc{\color{compc}compatible}}}
\newcommand{\vaultgold}{\ifblindvault independent blind labels\else second-annotator labels\fi}

\icmltitlerunning{Suggested Change Is a Separate Test for Memory-Retirement Gates}

\begin{document}
\raggedbottom

\twocolumn[
  \icmltitle{Suggested Change Is a Separate Test for Memory-Retirement Gates}

  \ifnum\arxivversion=1
\begin{icmlauthorlist}
  \icmlauthor{Xupeng Chen}{dg}
  \icmlauthor{Chenqian Le}{nyu}
\end{icmlauthorlist}
\icmlaffiliation{dg}{Dimension Gate}
\icmlaffiliation{nyu}{New York University}
\icmlcorrespondingauthor{Xupeng Chen}{xc1490@nyu.edu}
\icmlcorrespondingauthor{Chenqian Le}{cl6707@nyu.edu}
  \else
  \begin{icmlauthorlist}
    \icmlauthor{Anonymous Authors}{anon}
  \end{icmlauthorlist}
  \icmlaffiliation{anon}{Anonymous Institution}
  \icmlcorrespondingauthor{Anonymous Authors}{anonymous@example.com}
  \fi
  \icmlkeywords{long-term memory, memory update, evaluation, decision models, pre-registration}

  \vskip 0.3in
]

\printAffiliationsAndNotice{}

\begin{abstract}
Should a memory system delete a stored fact when a later remark only hints that it changed? Changed/unchanged labels
cannot separate a gate acting on established change from one acting on a hint. We write a contract separating the
two, relabel \nTest{} STALE test pairs and test pre-registered hypotheses on Jev, a typed decision model. Jev never
retired a claim on same-conversation non-changes, though its binary gate retired \fcKSTwo{} of \fcN{} change pairs gold
calls compatible. As a binary gate with an existing archive-change prompt it retires \preSTwo\% of claims whose change is
only suggested. A three-way gate written from the contract cuts this to \preSThree\% at a detection cost over our
pre-registered budget. It also differs in prompt and sees the earlier utterance; sent again with that prompt and input, a
yes/no output retires \mxYNPre\% and the three options \mxThreePre\%, so the prompt and input make the cut, not the
three options. The three-way rate comes from hints with
LLM-majority gold (\mNCSThree\%); on the \hNCN{} hints a human labelled, Jev's and AnyJev's three-way gates retire
\hNCSThree\% and \hNCSEight\% (exploratory). Some of these human labels are another author's, whose retire decisions
match an earlier AI pre-fill on every item. Memory builders should not
delete on the binary score of a gate unscored on suggested changes; our contract routes them to pending evidence
(untested end to end).
\end{abstract}

\section{Introduction}

\begin{figure*}[t]
\centering
\includegraphics[width=\textwidth]{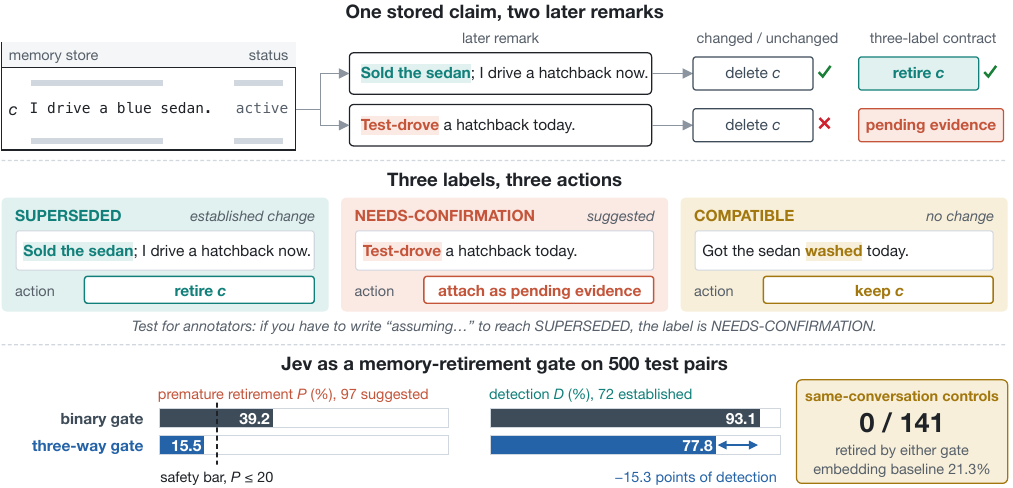}
\caption{\textbf{Jev retired none of \nWithinComp{} same-conversation non-changes, but as a binary gate it retires
\preSTwo\% of suggested changes.} \emph{Top:} read as changed or unchanged, both remarks delete the claim; the contract retires it only on the
established change. \emph{Middle:} the three labels (\cref{sec:task}) with remarks written for this figure.
\emph{Bottom:} Jev as a binary gate (retire if its score reaches $\tau{=}.5$; an existing archive-change prompt that sees
the claim and the later remark) and as a three-way gate (a prompt written from the contract that also sees the earlier
utterance; retire only on \SUP): $P$ on the \nTestNC{} \NC{} items with the pre-registered safety bar $P\le.20$, $D$ on the \nTestSup{} \SUP{}
items, and the \nWithinComp{} same-conversation controls, a subset of the \COMP{} items, of which neither gate retires
any ($0/\nWithinComp$; pooled $F$ in \cref{tab:test}). The three-way gate cuts $P$ from \preSTwo\% to \preSThree\% but
loses \hThreeLoss{} points of detection (\detSTwo\% to \detSThree\%). The bars thus compare two setups, not the
three-way output alone.}
\label{fig:overview}
\end{figure*}

An assistant that remembers its users must also forget. Suppose a user once said ``I drive a blue sedan.'' If they
later say ``Sold the sedan; I drive a hatchback now,'' the stored claim should be retired. If they say ``Test-drove a hatchback today,'' it should stay (\cref{fig:overview}). One error leaves stale facts in memory; the other deletes true facts on a hint. Recent memory systems and benchmarks
\citep{chhikara2025mem0,xu2025amem,wu2024longmemeval,hu2025memoryagentbench} evaluate the first direction well.

STALE \citep{chao2026stale} targets the update step directly: each of its 400 conflict scenarios pairs an early
utterance with a later one that, by construction, invalidates it, and it has no non-conflict pairs. It scores the
agent's answers, and its labels do not measure the second direction. Our pilot audit of STALE
pairs found that most retrieved ``changes'' needed an unstated assumption to count as one (a future plan, a different
group, two states that can coexist). A gate scored against those labels is rewarded for retiring on suggestion, and the reward is large. Only
\stalePairsEstablished{} of the \nChangeTest{} pairs STALE marks as changed in our test sample meet our task contract's
bar for an established change (\cref{sec:task,sec:data}). A gate that retired all \nChangeTest{} would be credited for
the other \the\numexpr\nChangeTest-\stalePairsEstablished\relax.

Fast typed decision models make the question practical. One call to Jev \citep{typesafe2026jev} returns a score in
$[0,1]$ (a \emph{Noul}) or a distribution over named options (a \emph{Choice}) \citep{typesafe2026docs}, priced by its vendor at \$0.042 per million input tokens \citep{typesafe2026jev}. Such models already serve as memory
controllers, judges and agent control layers \citep{jiang2026jevmem,li2026jevjudge,wu2026reflex}.
The closest work stops short of the retire decision. Jev-Mem \citep{jiang2026jevmem} keeps every observation and
defers obsolescence to a maintenance step that removes nothing. StateAuditor \citep{sun2026stateauditor} repairs an
agent's drafted response, not its store, and MemProbe \citep{ding2026memprobe} profiles whole memory systems. Whether such a model can be trusted to retire a
stored claim on its own has not been measured against labels that separate established from suggested change.

\begin{figure*}[t]
\centering
\includegraphics[width=\textwidth]{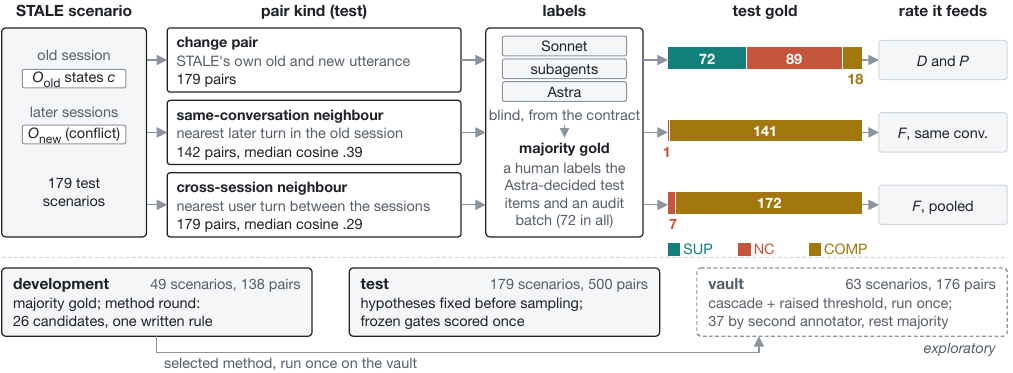}
\caption{\textbf{Each STALE scenario gives one change pair and two controls, and gold comes from three LLM annotators
with a human fallback.} \emph{Top:} pair kinds, labelling, test gold and the rate each kind mainly feeds ($P$ and $F$ pool over every pair of
their label). \emph{Bottom:} the three splits; on the vault the second annotator labelled the \nVaultHuman{} items that at least one
annotator marked as changed and the other items keep the three-LLM majority (\cref{sec:data}), so every vault result is
exploratory.}
\label{fig:protocol}
\end{figure*}

We therefore write a task contract with three labels and an action for each (\cref{fig:overview}). We relabel \nTest{} STALE
test pairs, same-conversation controls included, with three LLM annotators and a human fallback. We then
score Jev as a binary and a three-way gate against seven other gates, from a word-overlap rule to Mem0's own update step
(\cref{fig:protocol}).

Jev never retired a claim on \nWithinComp{} same-conversation
non-changes, but as a binary gate it retires \preSTwo\% of claims whose change is only suggested. Jev's three-way gate,
which also differs from the binary gate in prompt and input, lands within one item of where a raised threshold would (not pre-registered). With
that prompt and input, a yes/no output does as well as the three options (\cref{app:r9}). Of the gates in the main
comparison, no binary model gate both detects most established changes and meets the pre-registered safety bar on point estimates. Jev's and AnyJev's three-way gates meet
it on test only on point estimates whose intervals reach past the bar. The reasoning model (\preSFour\% on change pairs) was not run on the controls. Yet on the \hNCN{} hints a human labelled,
Jev's and AnyJev's three-way gates retire \hNCSThree\% and \hNCSEight\% (exploratory). Some of these labels are another
author's, whose retire decisions match an earlier AI pre-fill (\cref{sec:data}). Our contributions:
\begin{itemize}
\item \textbf{Contract and labels.} A task contract that splits ``changed'' into established (\SUP), suggested (\NC)
and no change (\COMP), with an action for each, and a relabelled STALE sample with natural same-conversation controls
(\cref{sec:task,sec:data}).
\item \textbf{Evaluation.} A pre-registered evaluation of Jev against lexical, embedding and reasoning-model
baselines, extended under a second written plan to an NLI cross-encoder, Mem0's update step and a second typed
decision model (\cref{sec:results}).
\end{itemize}

\section{Related work}

\paragraph{Memory update and forgetting.} Mem0 \citep{chhikara2025mem0} has an LLM choose add, update, delete or no-op
for each fact, so its delete branch is a retirement gate, scored only through downstream answers; A-MEM
\citep{xu2025amem} links and evolves notes. Benchmarks test whether an assistant answers with the latest state
\citep{wu2024longmemeval,maharana2024locomo,hu2025memoryagentbench}, STALE \citep{chao2026stale} isolates the update
step, and TANGLE \citep{yang2026tangle} scores conflicts with no single answer. StateMem \citep{fan2026statemem},
selective forgetting \citep{li2026forget} and memory-graph evolution \citep{jiang2026hage} study how state is kept. Supersede \citep{patel2026supersede} trains agents to keep a changed fact's current value, and
ConsistencyGate \citep{zhang2026consistencygate} admits a fact only when repeated queries support it. Concurrently,
\citet{deng2026stale} find that false merges silently remove current values. These works score whether the latest value
is recovered; we score whether an earlier one was deleted on sufficient evidence, Mem0's own update step included.

\paragraph{Fast typed decision models.} Jev returns a score or a distribution over named options from one call
\citep{typesafe2026jev,typesafe2026docs}. Jev-Mem \citep{jiang2026jevmem} uses it as a memory control plane without a
retirement operation, so it is not a baseline here. Others use it as a judge \citep{li2026jevjudge} or to control agents \citep{wu2026reflex}. Other work evaluates decision
models as text annotators \citep{ibrahim2026decision}, benchmarks Jev zero-shot on 37 public datasets
\citep{deusser2026jev}, and shows that its choices depend on option names \citep{sun2026typesafe}. AnyJev \citep{zhang2026anyjevreport,zhang2026anyjev} brings the interface to open-weight
models, removes position bias with option rotation and a label-free prior, and fits linear heads on intermediate layers. Its accuracy is reported on the public JevBench subset against that benchmark's released expected answers. We replicate the name effect on a new task at a smaller magnitude (\cref{sec:results}).

\paragraph{Evaluation with LLM labels.} LLMs serve as annotators \citep{gilardi2023chatgpt} and judges
\citep{zheng2023judging}. Judges favour their own family \citep{panickssery2024llm}, and a judge no more accurate
than the model it scores can at best halve the gold labels an evaluation needs \citep{dorner2024limits}. Our contribution here is narrow: a
pre-registered fallback that measured the problem (\astraAuditOver/\astraAuditN{} overturns) and forced human labels
where it mattered, and an identity for how much of a system's error a gold that shares it can hide.

\section{Task and estimands}\label{sec:task}

A \emph{pair} consists of a target claim $c$ stated by a user in utterance $O_{\text{old}}$ at time
$t_{\text{old}}$ and a later utterance $O_{\text{new}}$ at $t_{\text{new}} \ge t_{\text{old}}$. The label
$y \in \{\SUP, \allowbreak \NC, \allowbreak \COMP\}$ answers: as of $t_{\text{new}}$, what does $O_{\text{new}}$
establish about $c$? The label is \SUP{} only when five conditions hold without any unstated assumption. Subject, attribute and scope are
the same; $O_{\text{new}}$ asserts a current or completed state (not a plan, wish or question); and the new value is
incompatible with $c$. It is \NC{} when $O_{\text{new}}$ is evidence that $c$ may have changed but
reaching \SUP{} needs an assumption, and \COMP{} otherwise. Annotators get one test: \emph{if you have to write
``assuming\ldots'' to reach \SUP, the label is \NC.} A memory system may retire $c$ (and only $c$) on \SUP, should
attach $O_{\text{new}}$ as pending evidence on \NC, and should do nothing on \COMP.

A gate outputs a decision $r \in \{0,1\}$ (retire or keep). We estimate three conditional rates:
\begin{equation}
\begin{gathered}
D = \Pr(r{=}1 \mid y{=}\textsc{s}),\qquad
P = \Pr(r{=}1 \mid y{=}\textsc{n}),\\
F = \Pr(r{=}1 \mid y{=}\textsc{c}),
\end{gathered}
\end{equation}
where \textsc{s}, \textsc{n}, \textsc{c} abbreviate the three labels. These are detection, premature retirement and
false invalidation, each with a Wilson 95\% interval \citep{wilson1927}.

\paragraph{Hypotheses.} The pre-registered hypotheses were fixed before any sample was drawn, in a dated, hash-frozen
contract file kept outside version control. The within-session clause
of the safety bar and the scoring rules were added in dated amendments after sampling, before any human label or
system output had been read. \emph{Safety bar:} a gate may retire automatically only if $F \le .05$, both pooled and
on within-session controls, and $P \le .20$. \emph{Implicit gap:} on \SUP{} items, Jev's detection on implicit or
numeric changes is at least 20 points below explicit ones and the reasoning model beats it there (McNemar $p<.05$).
\emph{Three-way helps:} the three-way Choice has lower $P$ than the binary Noul at a detection loss of at most 10
points. The safety bar is a safety condition, not a utility criterion: a gate that never retires meets it (AnyJev's
binary gate passes at \detSSeven\% detection, \cref{sec:results}), so we read every pass with its detection $D$.
Binary gates retire at $\tau = .5$; three-way gates retire only on \SUP. Strata with $n<15$ are descriptive. All
$p$-values are exact McNemar tests on paired decisions \citep{mcnemar1947}, unadjusted.

\paragraph{Status of each analysis.} The gates added later, two retests and an output-format comparison (\cref{sec:data},
\cref{app:r8,app:r9}) have their own written plans, each committed to version control before any of its results existed. Everything else is exploratory:
\cref{sec:gold-source,sec:vault} and \cref{app:learned} throughout, and elsewhere each comparison that was not pre-registered says so.

\section{Data, gold labels and gates}\label{sec:data}

\paragraph{Pairs.} We drew from STALE's 400 conflict scenarios \citep{chao2026stale}, each an old-state and a
new-state utterance, excluding every scenario we had touched in pilot work. We split the rest by scenario into
development, test and a sealed vault (\uidDev, \uidTest{} and \uidVault{} scenarios yield pairs). Each scenario gives a
\emph{change pair} (STALE's old and new utterances) and two controls that STALE's construction implies should not be changes. The \emph{cross-session} neighbour is the user
turn between the two sessions most similar to $O_{\text{old}}$ under MiniLM (all-MiniLM-L6-v2;
\citealp{reimers2019sbert,wang2020minilm}); the \emph{within-session} neighbour is the most similar later turn in the
old session itself. The cross-session neighbour turned out to be mostly off-topic
(median cosine \csCos{} to the claim), so we added the within-session one (median \wsCos) before any annotation and
treat it as the primary false-invalidation control. We label \nTest{} test pairs (\nChangeTest{} change,
\nCrossTest{} cross-session, \nWithinTest{} within-session); the vault has \nVault{} (\nVaultChange, \nVaultCross,
\nVaultWithin).

\paragraph{Annotation.} Three LLM annotators labelled every pair blind to systems and to each other, from the contract text alone. They are
Claude Sonnet~5 through the vendor's local command-line client (the \emph{Sonnet annotator}), separate
Claude subagents (the \emph{subagent annotator}), and GPT-6 Astra (the \emph{Astra annotator}). Astra was added because the first two share a model family (\cref{app:method}). On change pairs the Sonnet annotator agrees poorly with the others
(Cohen's $\kappa$ \citep{cohen1960} \kappaOneTwo{} with the subagents, \kappaOneThree{} with Astra), the subagents and
Astra moderately (\kappaTwoThree), mostly disagreeing on the \SUP/\NC{} boundary (\cref{app:agreement}). Gold is the three-way majority, with human labels where a human
adjudicated, and the contract's eight dated addenda left the hypotheses unchanged (\cref{app:method}). One
author labelled a first audit batch of \nBatchOne{} items without opening the annotators' labels (the annotation page
logs no reveal; system outputs existed on disk but had not been summarised). A pre-registered fallback applied: that
author overturned the majority on \astraAuditOver/\astraAuditN{} items where the Astra annotator cast the deciding vote (threshold 15\%). Every such test item therefore went to human adjudication (dev items kept majority gold,
\cref{sec:vault}). On a random audit of the remaining items the author overturned \restAuditOver/\restAuditN{} (\cref{app:agreement}).
The 15\% fallback was written for the Astra-decided stratum, where the other two annotators had split; the random audit
has none, and its rate is reported as a limitation of majority gold (\cref{sec:limits}). Test gold is \nTestSup{} \SUP, \nTestNC{} \NC{} and \nTestComp{} \COMP, with
\nHumanTest{} items human-labelled. On the vault every item with any \SUP{} vote (\nVaultHuman) was labelled by the
second annotator described next.

\paragraph{Provenance of the \nReadj{} second-annotator labels.} The \nReadjTest{} remaining fallback items on test and the
\nVaultHuman{} vault items were first filled in on the adjudication pages by a coding agent (Codex) running
GPT-6 Astra, the model behind the Astra annotator. Both pages embed the three LLM annotators' labels behind a reveal
control; the agent reported having read them from the pages' source for the test items and not for the vault
items, a self-report we could not verify. We discarded the agent's labels. One author then relabelled the \nReadj{} items, and after that relabel had
been compared with the agent's labels, a second person, now one of the authors, labelled them on a page that
contains no model labels, attesting to working without AI tools or discussion. Our written rule required this annotator
not to have worked on the task; an author does not meet that condition. The rule that gold for these
items is the second annotator's first submission was written before that annotation; after the agreement with
the agent's labels was known, we first set the rule aside and then applied it by the authors' decision. The second
annotator agrees with the agent's three-way label on \labExtCodex/\nReadj{} items and with the author's relabel on
\labExtAuth/\nReadj; on the retire-or-keep decision that
every number in \cref{sec:vault} uses, the second annotator and the agent agree on \retExtCodex/\nReadj{}
(\retExtCodexVault/\nVaultHuman{} on the vault). Agreement with each LLM annotator and with the change-expression
field is in \cref{app:agreement}. The \nReadj{} labels were committed over \extSpanMin{} minutes, a median of \extMedianS~s after
each item was first shown. For scale, GPT-6 Astra run on the author's \nBatchOne{} first-batch items with the same written
instructions and no labels matched the author's three-way label on \astraHumanLabel/\nBatchOne{} and the retire
decision on \astraHumanRetire/\nBatchOne. We therefore report the \nReadj{} labels as a reading that two people
confirmed after a model fill, not as an independent human reading: the test verdicts do not depend on them
(\cref{sec:results}); the vault analysis does, and we label it exploratory.

\paragraph{STALE's labels.} All \nChangeTest{} test change pairs are ``changed'' in STALE; \stalePairsEstablished{}
of them are \SUP{} under the contract, and the rest are mostly \NC. This is partly definitional, since the contract's
same-attribute condition excludes STALE's Type~II propagation, and partly empirical (\cref{app:agreement}). We report
STALE's labels separately and do not replace them: STALE evaluates state resolution, including dependency-driven changes
in applicability; we ask the narrower question whether one stored claim may be deleted.

\paragraph{Gates.} \Cref{tab:test} gives each gate a short ID. The \emph{overlap rule} retires when
$O_{\text{new}}$ contains a change cue and its word overlap with $O_{\text{old}}$ passes a cutoff. The \emph{embedding
baseline} retires when the MiniLM cosine between $c$ and $O_{\text{new}}$ is at least .5. Both cutoffs were fixed
before any output was read. The \emph{binary gate} is Jev (\texttt{jev-1.13.0}) as a Noul with the wording of an
existing archive-change judge, unchanged: it sees $c$, $O_{\text{new}}$ and both times, not $O_{\text{old}}$, and its
prompt flags a claim for review, not deletion. Jev's training data are not disclosed, so STALE (public since 2026) may be
among them; Jev's rates could then reflect familiarity with the benchmark. The
\emph{three-way gate} is a Jev Choice that asks the contract's question, sees $O_{\text{old}}$ as well, and whose options
carry the contract's label names and definitions, developed on dev only. The two gates thus differ in output, prompt,
input and development. A \emph{neutral-name} variant
replaces the names with letters, and a \emph{swapped-name} variant gives each definition another label's name (the
\SUP{} definition is named \textsc{compatible}, and so on). The decision is always read from the definition chosen. The \emph{reasoning model} is Claude Sonnet~5 through the vendor's local command-line client
with the contract text, run on change pairs only. These gates were frozen, and their test outputs stayed sealed until
the first scoring. A separate written plan, fixed before any of their outputs existed, added four more. The \emph{NLI baseline} is an
off-the-shelf DeBERTa-v3-large cross-encoder \citep{he2021debertav3} trained on SNLI and MultiNLI
\citep{bowman2015snli,williams2018mnli} that retires when $O_{\text{new}}$ contradicts $c$ with probability at least .5.
\emph{Mem0's update step} runs Mem0 1.0.0's extraction and update prompts \citep{chhikara2025mem0,mem0repo} on
Qwen3-8B \citep{yang2025qwen3} with $c$ as the only stored memory and retires when it marks $c$ UPDATE or DELETE.
\emph{AnyJev's binary and three-way gates} put the same two questions, with the same inputs, word for word to a second typed decision
model, AnyJev \citep{zhang2026anyjevreport,zhang2026anyjev} on Qwen3-8B (setup in \cref{app:method}). Gold was
later revised twice, for the provenance reason above, by annotators who saw no system output; no system was re-run and
no verdict changed (\cref{sec:gold-source}). The first scoring's test aggregates also informed the method round
(\cref{sec:vault}).

\section{Results on test}\label{sec:results}

\begin{table*}[t]
\centering\small
\caption{\textbf{Test results} ($n{=}\nTest$). Rates in \%, Wilson 95\% intervals; false invalidation pooled over
\COMP{} items, within-session on its \nWithinComp{} \COMP{} controls. AUROC separates \SUP{} from the rest. Gate IDs:
S0 overlap rule, S1 embedding baseline, S2 binary gate, S3 three-way gate (S3n neutral names, S3x swapped names),
S4 reasoning model, S5 NLI baseline, S6 Mem0's update step, S7 and S8 AnyJev's binary and three-way gates.
$^\dagger$S4 ran on the \sFourCov{} change pairs only: its premature rate is over the \sFourNC{} \NC{} change pairs and its
false-invalidation rate over the \fcN{} \COMP{} change pairs, not the column denominators. $^\ddagger$Run later under
a separate written plan fixed before any of its outputs existed (\cref{app:method}); S6 counts every UPDATE or
DELETE event on $c$, including rewrites that leave $c$ unchanged.}
\label{tab:test}
\setlength{\tabcolsep}{4pt}
\begin{tabular}{@{}llllll@{}}
\toprule
System & Detection $\uparrow$ & Premature $\downarrow$ & False inv. $\downarrow$ & Within-sess. & AUROC \\
 & \scriptsize $n{=}72$ & \scriptsize $n{=}97$ & \scriptsize $n{=}331$ & & \\
\midrule
S0 Rule: $O_{\text{old}}$ overlap & 33.3 {\scriptsize [23.5, 44.8]} & 21.6 {\scriptsize [14.6, 30.8]} & 5.1 & 8.5 & -- \\
S1 MiniLM cosine & 19.4 {\scriptsize [12.0, 30.0]} & 5.2 {\scriptsize [2.2, 11.5]} & 9.7 & 21.3 & 0.542 \\
S2 Jev Noul, $\tau{=}.5$ & 93.1 {\scriptsize [84.8, 97.0]} & 39.2 {\scriptsize [30.1, 49.1]} & 1.2 & 0.0 & 0.963 \\
S3 Jev Choice (rubric names) & 77.8 {\scriptsize [66.9, 85.8]} & 15.5 {\scriptsize [9.6, 24.0]} & 0.9 & 0.0 & 0.957 \\
S3n \quad neutral names & 69.4 {\scriptsize [58.0, 78.9]} & 15.5 {\scriptsize [9.6, 24.0]} & 0.9 & 0.0 & 0.966 \\
S3x \quad swapped names & 51.4 {\scriptsize [40.1, 62.6]} & 6.2 {\scriptsize [2.9, 12.8]} & 0.3 & 0.0 & 0.962 \\
S4 Claude Sonnet 5 (change pairs)$^\dagger$ & 80.6 {\scriptsize [70.0, 88.0]} & 14.6 {\scriptsize [8.7, 23.4]} {\scriptsize ($n{=}89$)} & 11.1 {\scriptsize ($n{=}18$)} & -- & -- \\
\midrule
S5 NLI cross-encoder$^\ddagger$ & 80.6 {\scriptsize [70.0, 88.0]} & 36.1 {\scriptsize [27.2, 46.0]} & 19.3 & 11.3 & 0.856 \\
S6 Mem0 update step, Qwen3-8B$^\ddagger$ & 33.3 {\scriptsize [23.5, 44.8]} & 8.2 {\scriptsize [4.2, 15.4]} & 3.0 & 6.4 & -- \\
S7 AnyJev Noul, Qwen3-8B$^\ddagger$ & 6.9 {\scriptsize [3.0, 15.2]} & 0.0 {\scriptsize [0.0, 3.8]} & 0.0 & 0.0 & 0.823 \\
S8 AnyJev Choice, Qwen3-8B$^\ddagger$ & 45.8 {\scriptsize [34.8, 57.3]} & 12.4 {\scriptsize [7.2, 20.4]} & 1.2 & 0.7 & 0.877 \\
\bottomrule
\end{tabular}

\end{table*}

\begin{figure*}[t]
\centering
\includegraphics[width=\textwidth]{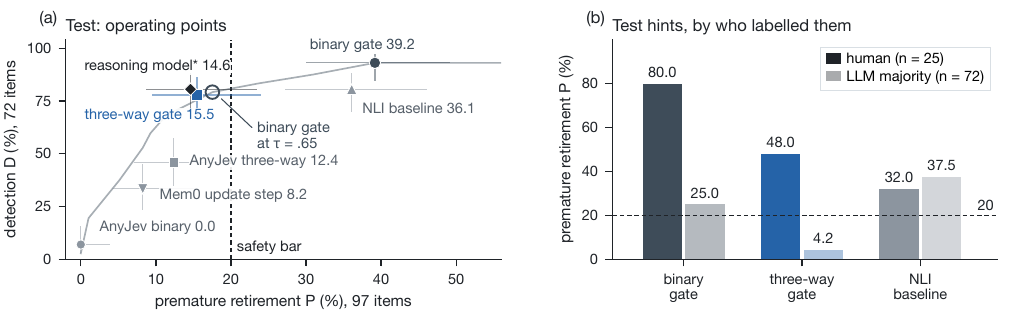}
\caption{\textbf{Jev's three-way gate sits on the binary gate's own threshold sweep, and its rate below the safety bar
comes from hints with LLM-majority gold.} The two gates differ in prompt and input as well as output
(\cref{sec:data}). (a) Test operating points, Wilson 95\% cross-hairs; the gray line sweeps the binary gate's
threshold from .05 to .95, and the ring marks $\tau{=}.65$, \detTauSixtyFive\%/\preTauSixtyFive\% (not pre-registered).
The dashed lines mark the safety bar $P\le.20$; the reasoning model's $P$ is over the \sFourNC{} \NC{} change pairs.
(b) $P$ on the \hNCN{} human-labelled and \mNCN{} LLM-majority \NC{} test items (exploratory); some of the human labels
are the second annotator's (\cref{sec:data}).}
\label{fig:frontier}
\end{figure*}

\subsection{Can any gate retire claims on its own?}

\paragraph{No Jev gate retired a same-conversation non-change.} No Jev variant retired any of the
\nNeighbourComp{} \COMP{} neighbour controls, including the \nWithinComp{} same-conversation ones, while the embedding
baseline retired \wsSOne\% of the within-session ones, so these controls do register false invalidation. The \fcN{}
change pairs that STALE calls changed but gold calls \COMP{} are harder. The binary gate retired \fcKSTwo{} of \fcN{}
change pairs, the three-way gate \fcKSThree{} and the reasoning model \fcKSFour, too few to separate the gates.

\begin{figure*}[t]
\centering
\includegraphics[width=\textwidth]{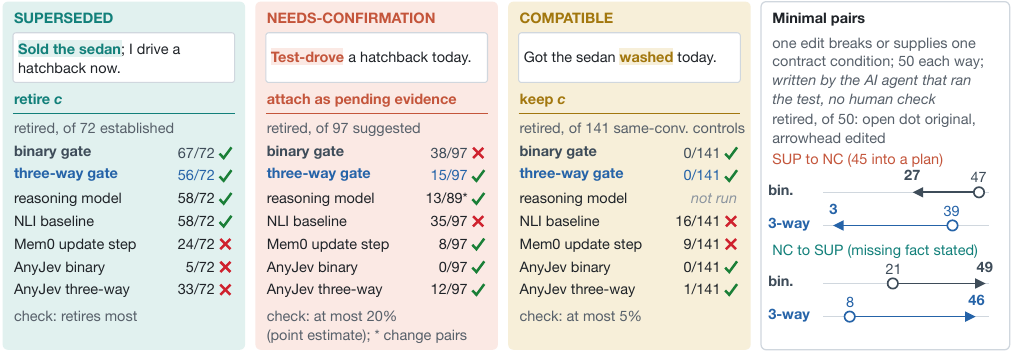}
\caption{\textbf{Of the gates drawn, only Jev's three-way gate both detects most established changes and passes the
safety bar, on point estimates; on minimal pairs the binary gate still retires \rAEditSTwo{} of \rPairsDir{} edited
members.} Cards: the remarks of \cref{fig:overview} (not STALE items), the contract's action, and how many test items of each
label each gate retires; a check marks retiring most \SUP{} items, at most 20\% of \NC{} items or at most 5\% of the
same-conversation controls (Mem0: event rule; \wsSSixCK/\nWithinComp{} counting content changes,
\cref{sec:results}). \emph{Right:} minimal pairs (\cref{app:r8}). The AI coding agent that ran this experiment
wrote the edits knowing the hypothesis, and no human, second author or checker reviewed them; the edited members have
no blind label, and most edits change exactly the feature the three-way gate's prompt names (plan or intent), so the
strip shows that it applies its stated rule to these texts, not how either gate reads natural data.}
\label{fig:cases}
\end{figure*}

\paragraph{As a binary gate, Jev retires on hints.} The binary gate fails the safety bar: it retires \preSTwo\% of
claims whose change is only suggested (\cref{tab:test}, \cref{fig:cases}).

\paragraph{The three-way gate passes the safety bar only on its point estimate.} The three-way gate retires
\preSThree\% (interval [\preSThreeLo, \preSThreeHi]) and passes on the point estimate, the reading the contract fixed.
The interval includes the .20 bar.

\paragraph{Retiring on a hint is not specific to Jev.} The NLI baseline fails the safety bar on both counts: it
retires \preSFive\% of suggested changes (interval [\preSFiveLo, \preSFiveHi]) and \wsSFive\% of the \nWithinComp{}
same-conversation controls (\wsSFiveK{} items).

\paragraph{The gates that pass the safety bar do so by retiring little.} Mem0's update step retires few suggested
changes (\preSSix\%) but detects only \detSSix\% of established ones, and under its pre-registered event rule it fails on
the same-conversation controls (\wsSSixK{} of \nWithinComp, \wsSSix\%), \wsSSixNoop{} of them UPDATE events that rewrite $c$
unchanged. Counting only events that delete $c$ or change its text (post hoc), it retires \wsSSixCK{} of these controls
and meets the safety bar on point estimates at \detSSixC\% detection. AnyJev passes with both questions, but by retiring little. Its binary gate retires no suggested change and detects only
\detSSeven\% of established ones; its three-way gate detects only \detSEight\% of
established changes at \preSEight\% premature retirement. AnyJev's three-way gate passes, as Jev's does, on the point
estimate: its interval [\preSEightLo, \preSEightHi] includes the .20 bar. On the \hNCN{} human-labelled \NC{} items,
some labelled by the second annotator (\cref{sec:data}), it retires \hNCSEightK/\hNCN{} (\hNCSEight\%; \cref{app:method}).

\paragraph{The implicit-gap hypothesis could not be tested} (\hTwoN{} qualifying \SUP{} items; \cref{app:limits}).

\subsection{What does the three-way setup change?}

\paragraph{The prompt and input make the cut, not the three options.} Going from the binary to the three-way gate cuts premature
retirement from \preSTwo\% to \preSThree\% at a \hThreeLoss-point detection cost (premature: McNemar \hThreeB:\hThreeC,
$p = \hThreeP$), over the 10-point budget, so the three-way hypothesis fails. The two setups also differ in prompt, in input (only
the three-way gate sees $O_{\text{old}}$) and in development (\cref{sec:data}). A later pre-registered comparison changes
only the output (\cref{app:r9}). Sent again with the three-way gate's prompt and input, a yes/no output retires \mxYNPre\% and
the three options \mxThreePre\%, so the output alone changes premature retirement by less than 10 points (McNemar
\mxMcA:\mxMcB, $p=\mxMcP$; difference \mxDiff{} points, 95\% interval [\mxDiffLo, \mxDiffHi]); detection is
\mxYNDetK{} against \mxThreeDetK{} of \nTestSup. The binary gate, re-called with its own prompt, retires \mxSTwoPreK{} of
\nTestNC.

\paragraph{Its rates are within one item of a raised threshold's, but its score carries more.} The two gates rank items about equally
well (AUROC \aucSTwo{} and \aucSThree), and the three-way gate sits on the binary gate's threshold sweep (not pre-registered; \cref{fig:frontier}a). In a pre-registered retest (\cref{app:r8}), at
nearly matched detection (\rDetTau{} and \rDetSThree{} of \nTestSup) both gates retire \rPreTau/\nTestNC{} suggested
changes yet decide differently on \the\numexpr\rMcA+\rMcB\relax{} of \the\numexpr\nTestSup+\nTestNC\relax{} items
(McNemar $p=\rMcP$, not a test of equivalence), and the three-way gate's score improves a logistic fit ($p=\rLRP$). On minimal pairs, edits written by
the AI agent that ran the retest, the three-way gate flips its decision on \rFlipSThree/\rPairs{} and the binary gate
on \rFlipSTwo{} (\cref{fig:cases}).

\paragraph{The reasoning model lands close to the three-way gate.} On the same change pairs the reasoning model is close
to the three-way gate (detection \detSFour\% vs.\ \detSThree\%, premature \preSFour\% vs.\ \preSThreeCh\% on the
\sFourNC{} \NC{} change pairs). It shares a model family and definitions with two of the annotators, which favours it. For AnyJev, the three-way hypothesis fails in reverse (\cref{app:method}).

\paragraph{Option names move the decision; AUROC changes little.} In the planned name variants (descriptive), neutral names
change \nameN\% of the three-way gate's decisions and swapped names \nameX\%, and detection falls from \detSThree\% to
\detSThreeX\%, while AUROC stays within .01 of the three-way gate's. A deployment that reads Jev's argmax inherits this
bias \citep{sun2026typesafe}; dividing out a label-free option prior does not remove it (exploratory, \cref{app:method}).

\subsection{Does the answer depend on who labelled the hints?}\label{sec:gold-source}

\paragraph{The verdicts hold under three readings of the gold.} With the \nReadjTest{} second-annotator test items scored
by the LLM majority instead, the binary gate retires \sensBOneSTwoPre\% and the three-way gate \sensBOneSThreePre\% of
\NC{} items ($n{=}\sensBOneNCN$). The three-way gate then loses \sensBOneLoss{} points of detection; with no human labels
at all, the figures are \sensNoSTwoPre\%, \sensNoSThreePre\% ($n{=}\sensNoNCN$) and \sensNoLoss{} points. Every safety-bar and
three-way verdict is unchanged under all three readings.

\paragraph{Where the gold came from changes the rate.} On the \hNCN{} human-labelled \NC{} items, \hNCAstra{} of which reached a human because the
Astra annotator cast the deciding vote between split annotators, the binary gate retires \hNCSTwo\% and the three-way
gate \hNCSThree\%. On the \mNCN{} \NC{} items with LLM-majority gold, they retire \mNCSTwo\% and \mNCSThree\%
(\cref{fig:frontier}b). Some of the human labels here are the second annotator's, whose retire decisions match the
discarded AI pre-fill on all \nReadj{} items they labelled, so these rates are exploratory (\cref{sec:data}).
The three-way gate's
test rate below the .20 bar comes from hints with LLM-majority gold (\mNCSThree\%).

\section{Limitations}\label{sec:limits}

Each human-labelled item has one adjudicator (one author for the first batch, the second annotator for the other \nReadj). The human labels are therefore one reading, not a consensus; their provenance is given in \cref{sec:data}.
Gold on the \the\numexpr\nTest-\nHumanTest\relax{} test items not sent to the
human is the LLM majority, and the random audit overturned \restAuditOver/\restAuditN{} of those. Our one held-out check
is exploratory, because every vault human label is the second annotator's and no independent blind relabelling was
done: there a cascade selected on dev retired \vpreMFour\% of suggested changes (\dpreMFour\% on dev), as many as a
raised threshold, and Jev's three-way gate retired \vpreSThreeV\% (\cref{sec:vault}). All Jev results are for one pinned version (\texttt{jev-1.13.0}), and every
result is on one English benchmark (STALE). The hosted service is also close to, not exactly, deterministic: identical
requests reproduce a score only approximately, and a retest changed \rtFlipK/\rtN{} binary-gate decisions. Every other
Jev decision here comes from one request. We did not analyse cost or latency, nor the end-to-end effect inside a memory
system. \Cref{app:limits} lists the remaining limitations.

\section{Conclusion}\label{sec:conclusion}

We asked whether a fast typed decision model can be trusted to retire stored claims. Scored against a contract that
separates established from suggested change, Jev retired none of \nWithinComp{} selected same-conversation controls (not
evidence of safety on natural update streams) but retires on hints. The binary score retires \preSTwo\% of
suggested changes. The three-way gate, a different prompt that also sees the earlier utterance, cuts this to
\preSThree\% and lands, for Jev, within one item of both rates of the binary gate at a raised threshold (not pre-registered). The
prompt and input make the cut: asked with them, a yes/no output retires \mxYNPre\%. Every other gate in the main comparison either retires on hints or rarely retires at all
(\cref{tab:test}); even three-way gates retire \hNCSThree\% (Jev) and \hNCSEight\% (AnyJev) of human-labelled suggested
changes (exploratory; some of these labels are the second annotator's, \cref{sec:data}). Memory systems should
treat a positive gate decision as evidence to attach, not as permission to delete, unless the decision model has been scored on suggested changes. Update benchmarks should label established and
suggested change separately, since STALE's ``changed'' covers both. Evaluations on LLM-majority gold should report, per \cref{prop:shared}, how many of the
system's errors the gold can see. No independent blind relabelling was done (\cref{app:r9}). Next steps are blind labels from an annotator who is not an author, for the \mNCN{} hints with LLM-majority gold and
the vault, and an end-to-end test on a second benchmark.

\section*{Impact Statement}
This paper evaluates whether a fast decision model can be trusted to retire stored facts about a user, a step
that any assistant with long-term memory must perform. The risk it measures is concrete: a gate that retires on
suggested rather than established change deletes true facts on a hint, and benchmark labels that do not separate
the two reward that behaviour. The conversational pairs come from a public benchmark; the only human participants
were two of the authors, who labelled text pairs from that benchmark. Users of memory-enabled assistants are the ones affected: premature deletion removes true facts about them, and
missed deletion keeps stale or retracted ones. The mitigation we recommend, holding suggested changes as pending
evidence, trades storage for safety and leaves deletion to an explicit user or rule.

\ifnum\arxivversion=1
\paragraph{Competing interests.} The authors have no affiliation with TypeSafe AI and received no funding from it; Jev was accessed through its public API.
\fi

\bibliographystyle{icml2026}
\bibliography{refs}

\appendix
\onecolumn
\raggedbottom
The appendices use the short IDs of \cref{tab:test} for the gates (S0--S8; M4 the cascade, M5 the raised threshold),
H1, H2 and H3 for the safety bar, the implicit gap and the three-way hypothesis of \cref{sec:task}, and A1, A2 and A3
for the Sonnet, subagent and Astra annotators of \cref{sec:data}.

\section{Reproducibility Statement and Use of AI Assistants}\label{app:statements}
\paragraph{Reproducibility statement.}
Every number, table and figure in this paper is generated by scripts from the saved annotator and system
outputs; a manifest of those input files (sizes and SHA-256 digests) and a claims checker that re-derives each
number from the per-item outputs will be released with the code. The task contract and its dated amendments, the
pre-registered hypotheses and selection rule, the item identifiers into STALE with every annotator's labels, the per-item
system outputs for test and vault, the gold files (majority votes and every human export) and the code will be released
upon publication; the two agent pre-fill exports, which the code repository alone cannot regenerate, ship with it.
Each Jev decision comes from a single request, apart from the \rtN{} S2 test items re-sent once (\cref{sec:limits}).

\paragraph{Use of AI assistants.}
AI assistants helped write some of the experiment code and check its correctness, helped collect public data,
helped check the correctness of citations, and helped polish the paper.

\section{Did the development estimate hold on held-out data?}\label{sec:vault}

This section is exploratory: every vault human label is the second annotator's (\cref{sec:data}).

\paragraph{Method round.} With test closed, we developed methods on dev under a written selection rule (highest
detection subject to $P \le .15$ and $F \le .05$). We then ran the winner and a re-thresholded binary gate once on the
vault (candidates in \cref{app:method}). The winner was
a \emph{cascade}: keep if the binary score is below .3, retire if it is at least .9, otherwise ask the reasoning
model. Its band, fixed before any reasoning-model output on dev, sent \vaultBand\% of vault items to the reasoning
model. On dev it reached \ddetMFour\% detection with \dpreMFour\% premature retirement, against
\ddetSTwoSeventy\%/\dpreSTwoSeventy\% for the binary gate at $\tau{=}.7$, the \emph{raised threshold}.

\begin{figure}[h]
\centering
\includegraphics[width=0.48\textwidth]{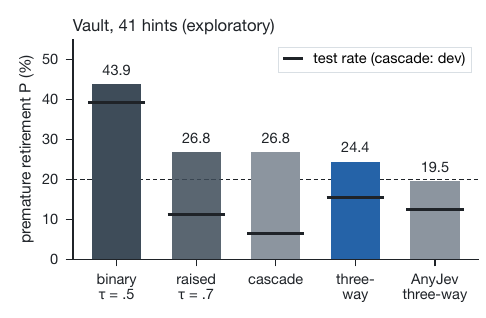}
\caption{Premature retirement on the held-out vault (\cref{sec:vault}). $P$ on the vault's
\nVaultNC{} \NC{} items (exploratory): \vNCHumK{} carry the second annotator's label (items that at least one annotator
marked as changed) and \vNCMajK{} the three-LLM majority; the dashed line marks the safety bar $P\le.20$; ticks: each gate's test rate (the cascade's: dev).}
\label{fig:vault}
\end{figure}

\paragraph{The development advantage did not replicate.} On the vault (\cref{tab:vault}, \cref{fig:vault}), the
cascade and the raised threshold both reach \vdetMFour\% detection with \vpreMFour\% premature retirement. That is far above the
cascade's dev estimate (premature rate \vpreMFour\% vs.\ \dpreMFour\% on dev) and the .15 target. Both improve on the binary gate at $\tau{=}.5$
(premature \vpreSTwoV\%; McNemar \vmcA:\vmcB, $p = \vmcP$), so this vault can register a gap of that size. On this
vault (\nVaultSup{} \SUP, \nVaultNC{} \NC) the reasoning model buys nothing that raising the threshold does not:
identical rates, and \the\numexpr\vDiffSup+\vDiffNC\relax{} discordant items split evenly; a smaller advantage would
not show. The three-way gate, run there later with its test prompt, detects as many established changes as the raised
threshold and retires one fewer of the \nVaultNC{} suggested changes
(\vdetSThreeV\%/\vpreSThreeV\% vs.\ the raised threshold's \vdetMFive\%/\vpreMFive\%; \cref{app:method}): on held-out
data, too, the three-way gate's rates are within one item of a raised threshold's.

\paragraph{Against majority gold the gap is smaller.} Scoring the same vault outputs against LLM-majority gold instead of the
\vaultgold{} on the \nVaultHuman{} items that carry them gives the cascade premature retirement \vMajPreMFourK/\vMajNCN{} (\vMajPreMFour\%) instead of
\vHumPreMFourK/\vHumNCN. Small-sample selection can also open such a gap: dev has only \nDevNC{} \NC{} items and the
winner was chosen among \nDevCand{} (method, threshold) candidates (a certified rule
would have refused the cascade, \cref{app:method}). The data neither apportion the gap between
gold and selection nor say which labels are correct. An identity says how much error a gold can hide relative to a
chosen reference reading.

\begin{proposition}\label{prop:shared}
Let $S$, $G$, $H \in \{0,1\}$ (retire or keep) be a system's decision, a gold decision and the true decision on an item. Then
\begin{equation*}
\begin{split}
\Pr(S \ne G) &= \Pr(S \ne H) + \Pr(G \ne H)\\
&\quad - 2\Pr(S \ne H,\ G \ne H).
\end{split}
\end{equation*}
Hence the error measured against $G$ understates the true error if and only if $\Pr(S\ne H, G\ne H) > \tfrac12 \Pr(G \ne H)$,
i.e., when the system shares over half of the gold's errors.
\end{proposition}

The proof is two lines (\cref{app:proof}); it needs binary decisions, so the three labels are collapsed to retire
(\SUP) or keep. \emph{Instance (\ifblindvault\else exploratory\fi).}
We treat the \ifblindvault\else second annotator's label\fi{} as $H$ (one reading, \cref{sec:data}\ifblindvault\else; it coincides with the discarded pre-fill on every retire
decision\fi). Then the \vBandN{} vault band items with such a label give $S{\ne}H$ on \shSH, $G{\ne}H$ on \shGH, and both on the same \shBoth{}
items, so $S{\ne}G$ on $\shSH + \shGH - 2\cdot\shBoth = \shSG$. Relative to that reading, majority gold shares \shBoth{}
of its \shGH{} errors with the escalation model (not distinguishable from half at this $n$) and cannot see them. It also charges the model with \the\numexpr\shGH-\shBoth\relax{} items on which the model agreed with the annotator; the net
understatement is \the\numexpr\shSH-\shSG\relax{} errors out of \shSH{} (\cref{app:agreement}). This data cannot say whether the \shBoth{} shared items are errors of the Claude-family models
or of a GPT-aligned label; only an annotator independent of both families can.

\paragraph{The vault in brief.} In short, this appendix holds a replication in which the method selected on development
data lost its advantage, and an identity for what a correlated gold hides, applied to the vault relative to one
annotator's reading. Both vault analyses are exploratory: they rest on another author's labels whose retire decisions
match an earlier AI pre-fill on all \nReadj{} items they cover (\cref{sec:data}). A reasoning-model cascade that looked best on \nDevNC{} development \NC{} items did not beat the raised threshold on a held-out vault. Against LLM-majority gold its vault premature rate is lower
(exploratory). Jev's three-way gate passes the safety bar on test only on its point estimate, and on the held-out vault it does not pass (\vpreSThreeV\%, interval
[\vpreSThreeVLo, \vpreSThreeVHi]). AnyJev's three-way gate passes on test, also only on its point estimate. On the vault it retires
\vpreSEightK/\nVaultNC{} (\vpreSEight\%), just under the bar. Jev's three-way gate's
test pass did not replicate on the vault; until independent vault labels exist, this and the rest
of the vault comparison stay exploratory. No independent blind relabelling was done (\cref{app:r9}).

\section{Proof of \cref{prop:shared}}\label{app:proof}
For binary values, $S \ne G$ holds exactly when one of $S$, $G$ differs from $H$ and the other does not, so
$\mathbf 1[S\ne G] = \mathbf 1[S\ne H] + \mathbf 1[G\ne H] - 2\,\mathbf 1[S\ne H]\,\mathbf 1[G\ne H]$ pointwise:
the right side is $0$ when neither or both differ from $H$ and $1$ when exactly one does. Taking expectations gives the
identity. Subtracting $\Pr(S\ne H)$, the measured error is smaller than the true error iff
$\Pr(G\ne H) - 2\Pr(S\ne H, G\ne H) < 0$. \qed

\section{Annotator agreement}\label{app:agreement}
\begin{figure}[h]
\centering
\includegraphics[width=0.48\textwidth]{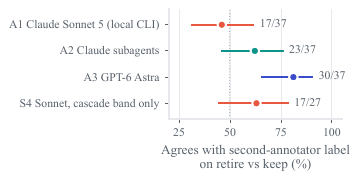}
\caption{Agreement with the \ifblindvault\else second annotator's label (exploratory)\fi{} (\cref{sec:data}) on retire vs.\ keep, vault items with any \SUP{} vote. Top three rows:
annotators on all \nVaultHuman{} items; bottom row: S4 on the \vBandN{} of them in M4's band. Wilson 95\% intervals.
Drawn from the per-item system outputs and gold files described in the Reproducibility Statement (\cref{app:statements}).}
\label{fig:agree}
\end{figure}
\begin{table}[h]
\centering\small
\caption{Confusion matrices on dev and test change pairs ($n{=}\nChangeAgree$); rows are the first annotator.}
\begin{tabular}{@{}lccccccccc@{}}
\toprule
 & \multicolumn{3}{c}{A1-A2 ($\kappa{=}0.28$, $n{=}228$)} & \multicolumn{3}{c}{A1-A3 ($\kappa{=}0.22$, $n{=}228$)} & \multicolumn{3}{c}{A2-A3 ($\kappa{=}0.60$, $n{=}228$)} \\
row $\backslash$ col & SUP & NC & COMP & SUP & NC & COMP & SUP & NC & COMP \\
\midrule
SUP & 56 & 17 & 0 & 40 & 33 & 0 & 71 & 35 & 0 \\
NC & 50 & 77 & 8 & 35 & 86 & 14 & 5 & 96 & 10 \\
COMP & 0 & 17 & 3 & 1 & 13 & 6 & 0 & 1 & 10 \\
\bottomrule
\end{tabular}

\end{table}
\begin{table}[h]
\centering\small
\caption{The second annotator's \nReadj{} labels (rows) against A1, A2, A3 and their majority (columns).}
\begin{tabular}{@{}lcccccccccccc@{}}
\toprule
 & \multicolumn{3}{c}{A1} & \multicolumn{3}{c}{A2} & \multicolumn{3}{c}{A3} & \multicolumn{3}{c}{majority} \\
second annotator $\backslash$ col & SUP & NC & COMP & SUP & NC & COMP & SUP & NC & COMP & SUP & NC & COMP \\
\midrule
SUP & 12 & 25 & 0 & 36 & 1 & 0 & 30 & 7 & 0 & 31 & 6 & 0 \\
NC & 15 & 11 & 0 & 18 & 8 & 0 & 3 & 23 & 0 & 9 & 17 & 0 \\
COMP & 3 & 2 & 0 & 3 & 2 & 0 & 0 & 4 & 1 & 1 & 4 & 0 \\
\bottomrule
\end{tabular}

\end{table}
\paragraph{STALE's labels by conflict type (\cref{sec:data}).} Of the \nTOne{} Type~I pairs, which change the same
attribute, \supTOne{} are \SUP; of the \nTTwo{} Type~II pairs, which change a related attribute whose consequence
invalidates the old claim, \supTTwo{} are \SUP{} and \ncTTwo{} are \NC. The contract's same-attribute condition excludes
Type~II propagation by construction, so ``most STALE changes are only suggested'' is in part a definitional statement
about Type~II; on Type~I it is an empirical one (\the\numexpr\nTOne-\supTOne\relax{} of \nTOne{} are not \SUP).

\paragraph{The random audit (\cref{sec:data}).} Of the \restAuditOver/\restAuditN{} randomly audited items the author
overturned, only \restAuditRetire, a unanimous \SUP{} change pair, changes a retire decision; \restAuditCtrl{} move
controls from \COMP{} to \NC{} and \restAuditToComp{} a change pair from \NC{} to \COMP.

\paragraph{The \nReadj{} second-annotator labels in full (\cref{sec:data}).} Requiring the change-expression field to
match as well as the three-way label, the second annotator agrees with the agent on \jointExtCodex/\nReadj{} and with
the author's relabel on \jointExtAuth/\nReadj. The author's relabel agrees with the agent on
\labAuthCodex/\nReadj{} (\jointAuthCodex/\nReadj{} with change expression); on the retire decision the second
annotator agrees with the author's relabel on \retExtAuth/\nReadj. The second
annotator agrees with A1, A2, A3 and their majority on \extAOne, \extATwo, \extAThree{} and \extMaj{} of the
\nReadj{}, and the three annotators agree pairwise on \pwOneTwo, \pwOneThree{} and \pwTwoThree{} of them.
On all \nBand{} band items the escalation model (M4's reasoning model) agrees with A2 on \bandAgreeTwo, with the
majority on \bandAgreeMaj, with A3 on \bandAgreeThree{} and with A1 on \bandAgreeOne; with the
\ifblindvault\else second annotator\fi{} it agrees on \vBandAgree/\vBandN{} (\cref{fig:agree}); of its \shSH{} disagreements
with that annotator, \shOver{} are retirements the annotator would have kept. We did not test why it tracks A2 so closely.
Detection by STALE conflict type (pre-registered stratum; Type~II is descriptive, $n{=}\supTTwo<15$): S2
\detSTwoTOne\% on the \supTOne{} Type~I \SUP{} items and \detSTwoTTwo\% on the \supTTwo{} Type~II ones; S3
\detSThreeTOne\% and \detSThreeTTwo\%.

\section{Method-development round and model identifiers}\label{app:method}
\begin{table}[h]
\centering\small
\caption{Dev (majority gold, selection) and vault (\vaultgold{} on all \nVaultHuman{} items with any \SUP{} vote and
the three-LLM majority on the other items,
\ifblindvault\else exploratory, \cref{sec:data}\fi). Rates in \%, Wilson 95\% intervals. Dev gold is the LLM
majority; \nDevAstra{} dev items (\nDevAstraSup{} of the \nDevSup{} \SUP, \nDevAstraNC{} of the \nDevNC{} \NC,
\nDevAstraBand{} of the \devBandN{} band items) were decided by the Astra annotator's vote and never human-checked,
because the fallback was scoped to test. M5: the raised threshold; M4: the cascade.}
\label{tab:vault}
\begin{tabular}{@{}lllll@{}}
\toprule
System (selected on dev) & Dev det. / prem. & Vault detection & Vault premature & Vault false inv. \\
 & \scriptsize $n{=}18/31$ & \scriptsize $n{=}20$ & \scriptsize $n{=}41$ & \scriptsize $n{=}115$ \\
\midrule
S2 Jev Noul, $\tau{=}.5$ & 94.4/38.7 & 100.0 {\scriptsize [83.9, 100.0]} & 43.9 {\scriptsize [29.9, 59.0]} & 0.9 \\
M5 Jev Noul, $\tau{=}.7$ & 77.8/9.7 & 95.0 {\scriptsize [76.4, 99.1]} & 26.8 {\scriptsize [15.7, 41.9]} & 0.9 \\
M4 Noul + Sonnet on $[.3,.9)$ & 94.4/6.5 & 95.0 {\scriptsize [76.4, 99.1]} & 26.8 {\scriptsize [15.7, 41.9]} & 0.9 \\
\bottomrule
\end{tabular}

\end{table}
Candidates, written before running: M1, five Nouls, one per contract condition, retire iff all exceed $\tau$; M2,
S2 AND M1; M3, mean of the Jev probabilities; M5, S2 re-thresholded; $\tau \in \{.5,.6,.7,.8\}$. M1 retired
almost nothing on dev because the scope question included ``time frame'', which every change violates.
Post-hoc variants using only the attribute and incompatibility questions (M1b, M2b, M6) did no better than M5. The
cascade M4 and its band were written down before any reasoning-model output on dev existed. All dev numbers use
majority gold; the \nDevCand{} candidates and the selection rule are replayed from the per-item outputs by the
released code.

\paragraph{Cascade and raised threshold on the vault.} They differ on \vDiffSupEach{} \SUP{} and \vDiffNCEach{} \NC{}
items in each direction.

\paragraph{Contract amendments.} The contract was amended in eight dated addenda, all released with it; the last
three, written after human labels existed, scoped the human fallback to test and clarified the second batch's
instructions, fixed the development round and added the second annotator, and none of them changed the hypotheses or
their thresholds.

\paragraph{Gates added under the second plan.} The NLI baseline S5 was not tuned. Mem0's update step S6 used greedy
decoding and showed $c$ as the only stored memory, so that retrieval cannot miss it. S7 and S8 use AnyJev's label-free
calibration (option rotation, option prior estimated on dev).

\paragraph{Model identifiers.} A1 and S4: Claude Sonnet~5 (\texttt{claude-sonnet-5}) through the vendor's local
command-line client, one pair per request. A2: separate Claude subagents of the same family. A3:
\texttt{gpt-6-astra} through a third-party API relay at reasoning effort \emph{medium} (every response reports \texttt{gpt-6-astra}); the coding agent that pre-filled the
\nReadj{} items ran the same model at effort \emph{high}. S2, S3, S3n, S3x: \texttt{jev-1.13.0}. Dev and test were
annotated and run on 23--24 September 2026, the vault on 27 September; the second annotator labelled on 28 September.
S3, S3n and S3x were run on the vault, and S2 re-sent on \rtN{} test items, on 14 October. S5:
\texttt{cross-encoder/nli-deberta-v3-large}, float32 on CPU, premise $O_{\text{new}}$ and hypothesis $c$, run on 14 October.
S6: \texttt{Qwen/Qwen3-8B} without thinking, served with vLLM in bfloat16, run on 14 October. S7, S8: AnyJev 0.2.0
with \texttt{Qwen/Qwen3-8B} in float32 through its Hugging Face backend, run on 15 October.

\paragraph{Dividing out an option prior does not remove the name effect (exploratory).} For open-weight models,
position bias in typed decisions is removed by rotating options and dividing out a prior estimated without
labels \citep{zhang2026anyjevreport,zhang2026anyjev}. The same label-free division applied to Jev (mean option probabilities estimated on dev, divided out on test)
leaves the name effect: the priors barely differ across namings (mean $P(\SUP)$ \pdPriorSupS{} for S3, \pdPriorSupX{}
for S3x), after division S3x still disagrees with S3 on \pdDivDisX\% of items (\pdRawDisX\% before; argmax
decisions), and its detection is \pdDivDetX\% against S3's \pdDivDetS\%. The effect varies by item rather than
acting as a per-option offset, which a correction built for position bias cannot reach.

\paragraph{Certified selection.} A selection rule that certifies its target would
have refused the cascade. The one-sided 95\% Clopper--Pearson bound \citep{clopper1934} on its dev premature retirement
(\cpKMFour/\nDevNC) is \cpUpMFour\%, above the .15 target; by coincidence this is the same figure as the cascade's majority-gold
vault rate (\cref{sec:vault}). Certifying .15 with \cpKMFour{} errors needs at least \cpNeedN{} \NC{} items. AnyJev bounds its early-stopping rule with the same kind of one-sided bound for the same
reason \citep{zhang2026anyjevreport}.

\paragraph{Retest of the hosted service.} Re-sent \rtDays{} days later, the binary gate's requests for \rtN{} test items
changed \rtFlipK/\rtN{} decisions at $\tau{=}.5$, all on \NC{} items, with no score moving by more than \rtMaxAbs. Our
pre-registered rule labels this as affecting the verdicts, because shifts that size could, in the worst case, close the
three-way hypothesis's margin of \rtHThreeMargin{} \SUP{} items; at the observed rate none changes.

\paragraph{The NLI baseline in more detail (exploratory beyond its H1 verdict).} On \NC{} items S2 and S5 retire
different claims: \ncSTwoOnlyFive{} are retired by S2 only and \ncFiveOnlyTwo{} by S5 only. Unlike Jev's, S5's premature
retirement does not concentrate on the human-labelled items (\hNCSFive\% on those, \mNCSFive\% on the LLM-majority
ones; \cref{sec:results}). On the vault S5 reaches \vdetSFive\% detection with \vpreSFive\% premature retirement.
With premise and hypothesis swapped, a variant the plan listed as exploratory, the cross-encoder meets H1 on test
(detection \detSFiveR\%, premature \preSFiveR\%, false invalidation \fiSFiveR\% pooled and \wsSFiveR\% within-session),
trading detection for premature retirement much as S3x does, and fails it on the vault (\vpreSFiveR\% premature).

\paragraph{AnyJev in more detail (exploratory beyond its H1 and H3 verdicts).} AnyJev's three-way gate retires \ajHThreeB{}
suggested changes that its binary gate keeps and none the other way (McNemar $p = \ajHThreeP$). Unlike Jev's, AnyJev's
three-way gate sits above its binary gate's threshold sweep (not pre-registered): at no more premature retirement than
the three-way gate, the binary gate detects at most \swSSevenD\%. On the vault S7 detects \vdetSSeven\%
with \vpreSSeven\% premature retirement and S8 \vdetSEight\% with \vpreSEight\%; both meet H1 there. Without AnyJev's
calibration (the canonical option order only), S8 detects \detSEightRaw\% but retires \preSEightRaw\% of suggested
changes and \wsSEightRaw\% of same-conversation controls, failing H1; with the option prior estimated on the scored
split instead of dev, both gates still meet H1. S8's AUROC is \aucSEight{} against \aucSSeven{} for S7.

\paragraph{Mem0's update step in more detail (exploratory beyond its H1 verdict).} On the \nTest{} test
items the extraction step returned no fact for \noCallSSix, so Mem0 made no update call and kept $c$. Of the
\updSSixN{} items on which the update step marked $c$ UPDATE, \updSSixSame{} rewrote $c$ with its own text, a change
that leaves the store as it was but counts as a retirement under the written rule. Counting only DELETE events and
UPDATEs that change $c$'s text (post hoc; texts compared as in the run, with the same result when case and punctuation
are ignored), S6 detects \detSSixC\% with \preSSixC\% premature retirement, retires \fiSSixC\% of \COMP{} items pooled
and \wsSSixCK{} of the \nWithinComp{} same-conversation controls, and meets H1 on test; on the vault it detects
\vdetSSixC\% with \vpreSSixC\% and retires \vwsSSixCK{} of \vwsN{} controls. This reading still counts an UPDATE that
rewrites $c$ into a compatible claim as a retirement, so it lies between the DELETE-only and the event rates. Counting DELETE only, a variant the
plan listed as exploratory, S6 meets H1 on test at \detSSixDel\% detection and \preSSixDel\% premature retirement. On
the vault S6 reaches \vdetSSix\% detection with \vpreSSix\% premature retirement and retires \vwsSSixK{} of \vwsN{}
same-conversation controls. With Qwen2.5-32B-Instruct in place of Qwen3-8B (exploratory) the same pipeline detects \detSSixX\% with
\preSSixX\% premature retirement and retires \wsSSixXK{} of the \nWithinComp{} same-conversation controls
(\wsSSixX\%), \wsSSixXNoop{} of them unchanged rewrites, so it too fails H1 only on those controls; counting content
changes only, it still retires \wsSSixXCK{} of them (\wsSSixXC\%).

\paragraph{Three-way gate on the vault (exploratory).} After the method round, S3, S3n and S3x were run once on all
\nVault{} vault items with their frozen test prompts and decision rule, under a written plan, and scored against the
vault gold (the second annotator's labels on the \nVaultHuman{} items that at least one annotator marked as changed, the
three-LLM majority on the rest). S3 reaches \vdetSThreeV\% detection with \vpreSThreeV\% premature retirement
([\vpreSThreeVLo, \vpreSThreeVHi]) and fails H1; S3n fails it too (\vpreSThreeNV\%), and S3x passes it
(\vpreSThreeXV\%) at \vdetSThreeXV\% detection, the same trade as on test. Against S2 at $\tau{=}.5$ on the vault, S3
meets the H3 criteria, unlike on test: on \NC{} items McNemar \vhThreeA:\vhThreeB, $p = \vhThreeP$, and a detection
loss of \vhThreeLoss{} points. We do not read this as support for H3. With \nVaultSup{} \SUP{} items one item moves the
loss by \vhThreeStep{} points; one discordant item fewer gives $p = \vhThreeOneLessP$ and two fewer
$p = \vhThreeTwoLessP$; and the S2 outputs it is compared with were obtained \vSThreeGapDays{} days earlier, a gap
over which re-sent S2 requests changed \rtFlipPct\% of decisions (\cref{sec:limits}).

\section{Do learned gates inherit the teacher's optimism? (exploratory)}\label{app:learned}
S7 and S8 use AnyJev without labels. AnyJev can also fit a linear head on a model's intermediate layers from labelled
examples \citep{zhang2026anyjevreport,zhang2026anyjev}, and the labels a builder is likely to have are LLM labels. A
separate study asked whether such a head inherits its teacher's premature retirements, under its own written plan, frozen
before any model output existed. It is exploratory here: no verdict in \cref{sec:results} depends on it.

\paragraph{Design.} The teacher is the LLM majority of A1--A3 (\cref{sec:data}). The heads are AnyJev~0.2.0 linear heads
on Qwen3-8B states with the library's defaults (candidate layers at 50--100\% depth; LDA or ridge chosen by five-fold
out-of-fold likelihood), asked S8's three-way question (retire on \SUP) or S7's binary one (retire if
$P(\text{Yes})\ge.5$). Unless stated otherwise they were fitted with the teacher's labels on \sbPool{} pool items: every
annotated item outside the evaluation set that shares no STALE scenario with it. The evaluation set is the \sbN{} items a human labelled
(\sbNTest{} on test, \sbNVault{} on the vault; \sbSup{} \SUP, \sbNC{} \NC, \sbComp{} \COMP), scored against this paper's
human labels; nothing reported here was fitted or selected on an evaluation item, label or statistic. Before these items were scored,
the plan's gates passed on the pool: out-of-fold three-way accuracy \sbGOneAcc{} against a majority-class rate of
\sbGOneBase, \SUP{} recall \sbGTwo, and \sbGFourK/\sbGFourN{} states re-extracted in a separate process gave the same
decision. The plan specified bfloat16, in which that last gate failed (\sbGFourKBf/\sbGFourN); a rule written before any
human label entered scoring made a float32 run primary and kept bfloat16 as a sensitivity run.

\begin{table}[h]
\centering\small
\caption{Learned gates on the \sbN{} human-labelled items (exploratory, \cref{app:learned}): AnyJev~0.2.0 on Qwen3-8B
in float32, linear heads fitted on \sbPool{} pool items with the labels in parentheses. Rates in \%, Wilson 95\%
intervals. Dose rows: a fraction $\rho$ of the \sbDoseN{} pool change pairs the teacher calls \NC{} relabelled \SUP{}
before fitting; means over \sbDoseDraws{} relabelling draws.}
\label{tab:learned}
\begin{tabular}{@{}llll@{}}
\toprule
Gate (training labels) & Detection $\uparrow$ & Premature $\downarrow$ & False inv. $\downarrow$ \\
 & \scriptsize $n{=}48$ & \scriptsize $n{=}41$ & \scriptsize $n{=}20$ \\
\midrule
LLM majority (the teacher) & 79.2 {\scriptsize [65.7, 88.3]} & 31.7 {\scriptsize [19.6, 47.0]} & 5.0 \\
AnyJev three-way, zero-shot (prior from the pool) & 35.4 {\scriptsize [23.4, 49.6]} & 24.4 {\scriptsize [13.8, 39.3]} & 5.0 \\
Linear head, three-way (majority) & 45.8 {\scriptsize [32.6, 59.7]} & 34.1 {\scriptsize [21.6, 49.5]} & 5.0 \\
\quad at the pool-certified $\tau^*{=}.158$ & 60.4 {\scriptsize [46.3, 73.0]} & 41.5 {\scriptsize [27.8, 56.6]} & 5.0 \\
Linear head, binary (majority) & 58.3 {\scriptsize [44.3, 71.2]} & 41.5 {\scriptsize [27.8, 56.6]} & 0.0 \\
Linear head, binary (STALE's labels) & 100.0 {\scriptsize [92.6, 100.0]} & 92.7 {\scriptsize [80.6, 97.5]} & 35.0 \\
\midrule
Three-way head, dose $\rho{=}.25$ & 62.9 & 48.3 & 12.0 \\
Three-way head, dose $\rho{=}.5$ & 73.3 & 61.0 & 16.0 \\
Three-way head, dose $\rho{=}1$ & 100.0 & 85.4 & 35.0 \\
\bottomrule
\end{tabular}

\end{table}
\paragraph{Results (\cref{tab:learned}).} \emph{Instrument check.} The binary head fitted on STALE's construction labels
(every change pair retired) retires \sbStaleK/\sbNC{} suggested changes and the same head on majority labels
\sbMajQNK/\sbNC{} (McNemar \sbStaleDisc:\sbStaleDiscB, one-sided $p=\sbStaleP$): the head follows its labels on this
boundary, so a null result below is informative. \emph{Rate.} The three-way head on majority labels retires
\sbHeadPK/\sbNC{} suggested changes (\sbHeadP\%, [\sbHeadPLo, \sbHeadPHi]) to the teacher's \sbMajPK/\sbNC{}
(\sbMajP\%), at least the teacher's rate minus 10 points, which the plan reads as inheriting it. \emph{Items.} It does
not reliably retire the same ones: it retires \sbShareA{} of the \sbShareNOne{} suggested changes the teacher retires and
\sbShareB{} of the \sbShareNTwo{} it keeps (Fisher one-sided $p=\sbFisherP$; Holm-adjusted \sbFisherHolm{} over the plan's
three confirmatory tests). The plan's verdict is mixed: the head inherits the teacher's rate, not its items.
\emph{Dose.} Relabelling as \SUP{} a fraction $\rho$ of the \sbDoseN{} pool change pairs the teacher calls \NC{} raises
mean premature retirement from \sbDosePZero\% ($\rho=0$) through \sbDosePQuarter\% and \sbDosePHalf\% to \sbDosePOne\%
($\rho=1$; \sbDoseDraws{} draws per $\rho$), with no dip, and detection from \sbDoseDZero\% to \sbDoseDOne\%.
\emph{Labels against none.} Compared with AnyJev's zero-shot three-way gate with the option prior estimated on the pool
(not S8, whose prior comes from dev), the head differs on neither rate (premature \sbLTwoOnlyP:\sbLZeroOnlyP,
$p=\sbLZeroLTwoPP$; detection \sbLTwoOnlyD:\sbLZeroOnlyD, $p=\sbLZeroLTwoDP$). In bfloat16 the head detects more
(\sbBfLTwoOnlyD:\sbBfLZeroOnlyD, $p=\sbBfLZeroLTwoDP$), so this contrast is not robust to numerical precision; every other
verdict here is the same in both runs. \emph{Certified threshold.} The lowest threshold on the head's out-of-fold pool
$P(\SUP)$ whose one-sided 95\% Clopper--Pearson bound on premature retirement among the pool's majority-\NC{} items is at
most .20 (and among its majority-\COMP{} items at most .05) is $\tau^*=\sbTau$ (bound \sbCertNC\%,
\sbCertNCK/\sbCertNCN). On the \sbN{} items it retires \sbFivePK/\sbNC{} suggested changes (\sbFiveP\%) at \sbFiveD\%
detection, and only \sbFiveOverlap{} of those \sbFivePK{} are the teacher's own retirements, which the plan reads as
distribution shift (these items are harder than the pool) rather than as the certificate inheriting the teacher's
optimism. Either way, a bound certified on LLM-labelled pool items did not hold on the human-labelled ones.

\paragraph{\Cref{prop:shared} on these items.} With the head as $S$, the majority as $G$ and the human label as $H$
(retire if \SUP) on all \sbN{} items, $S{\ne}H$ on \sbSH, $G{\ne}H$ on \sbGH{} and both on \sbBoth, so $S{\ne}G$ on
$\sbSH + \sbGH - 2\cdot\sbBoth = \sbSG$. The head shares exactly half of the majority's errors, the boundary case of the
proposition, so its error against majority gold equals its error against these labels: majority gold hides \sbBoth{} of
the head's errors and charges it with \the\numexpr\sbGH-\sbBoth\relax{} items on which it agreed with the human. Inheriting
a teacher's rate without its items therefore does not by itself make majority gold understate a system's error; the
escalation model in \cref{sec:vault}, which shares \shBoth{} of the majority's \shGH{} errors on the band, is a case that does.

\paragraph{Caveats.} Apart from a random audit of \restAuditN{} test items, the \sbN{} items reached a human because the
annotators split or, on the vault, because one voted \SUP, so they are a contested subset and no rate here is a population
rate. On the pool each annotator calls \SUP{} at most \sbPoolLoneK{} of the \sbPoolLoneN{} or more items the other two call
\NC, so a head fitted there sees little of a teacher's optimism, which biases toward finding no inheritance through
under-exposure (the instrument check and the dose response are there to separate the two). Of the \sbN{} labels,
\nReadj{} are the second annotator's, with the provenance given in \cref{sec:data} and \cref{app:agreement}; on the \sbFirstN{} first-batch items, which one
author labelled, the head retires \sbFirstHeadK/\sbFirstNC{} suggested changes and the teacher \sbFirstMajK/\sbFirstNC,
the same sign as on all \sbN. With \sbNC{} \NC{} items only differences of about 20 points are detectable. The study used
one model (Qwen3-8B); the plan's replications on Qwen3-32B and Qwen3-4B have not been run.

\section{Further limitations}\label{app:limits}
\paragraph{The implicit-gap hypothesis.} Only \hTwoN{} \SUP{} items in its test set (human-labelled
or unanimous across the three annotators; \hTwoHuman{} and \hTwoUnan{} of the \hTwoN) are implicit or numeric.
Descriptively the direction is reversed (binary gate \hTwoSTwo\%, reasoning model \hTwoSFour\%). The contract sends most
implicit changes to \NC{} by definition, so an implicit-change test needs a different sample.

\paragraph{Other limitations.} Of the test items with LLM-majority gold (\cref{sec:limits}), the controls, on which no annotator ever voted \SUP, are
the least affected by its errors.
Instructions to the human were clarified between two batches, so human-label statistics should be read per batch; gold
uses the first-submission labels throughout, and the labels the human revised after discussing batch-one items with an
assistant leave every verdict unchanged. The implicit-gap hypothesis had \hTwoN{} items. For AnyJev the three-way
decision is not a threshold shift of the binary one, so the threshold-frontier finding is Jev's; whether the name effect
or any finding holds for other typed decision models or update benchmarks is untested; a later Jev version is a
different instrument. The \rtFlipK/\rtN{} retest flips are ones our pre-registered rule labels as affecting the
verdicts, though at that rate none changes (\cref{app:method}). The vault is spent and no unused STALE scenarios remain,
so the observation that GPT-6 Astra agreed most with the vault \ifblindvault\else second-annotator labels\fi{}
(\vAgreeThree/\nVaultHuman)\ifblindvault\else, labels that match the Astra-driven agent's fill on
\retExtCodexVault/\nVaultHuman{} retire decisions\fi, needs new data. The contract also listed cost and latency, which
we did not analyse. Mem0's update step is scored with one open model and perfect retrieval, not Mem0 as deployed.

\section{Threshold shift and minimal pairs (pre-registered after the main results)}\label{app:r8}
Two tests were written down (PREREG\_R8) after the results above, before any number in this section was computed; the
minimal-pair items were committed before they were scored. S5 and S8 were not run here (no GPU was available).

\paragraph{Is S3 a threshold shift of S2?} On the \nTestSup{} \SUP{} and \nTestNC{} \NC{} test items (\COMP{} excluded) we
chose the S2 threshold whose detection matches S3's, $\tau^*=\rTau$ (ties \rTauLo--\rTau; S2 \rDetTau/\nTestSup, S3
\rDetSThree/\nTestSup). There both gates retire \rPreTau/\nTestNC{} suggested changes and disagree on \rMcA:\rMcB{} items
(exact McNemar $p=\rMcP$; \rMcSensP{} at $\tau=\rTauLo$). Their AUROCs on \SUP{} vs.\ \NC{} are \rAucSTwo{} and \rAucSThree{}
(DeLong difference \rAucDiff, 95\% interval [\rAucLo, \rAucHi], $p=\rAucP$). But in a logistic model of \SUP{} vs.\ \NC,
S3's $P(\SUP)$ added to S2's score improves the fit (LR $=\rLR$, $p=\rLRP$; entering S2 as a logit does not change this,
and S2 added to S3 does not help, $p=\rLRRevP$; both not pre-registered). The rule required all three tests to pass,
so ``threshold shift'' is not supported: at approximately matched detection the two gates retire the same number of
\NC{} items but decide differently on \the\numexpr\rMcA+\rMcB\relax{} of the \the\numexpr\nTestSup+\nTestNC\relax{}, and a
non-significant McNemar test does not show that they decide alike; S3's score carries information about \SUP{} vs.\
\NC{} that S2's does not. On the vault (exploratory) none of the three tests rejects
(McNemar \rvMcA:\rvMcB; DeLong [\rvAucLo, \rvAucHi]; LR $p=\rvLRP$). S3's prompt was developed on dev and S2's was not, and S3 also sees $O_{\text{old}}$, so none of these tests isolates the
output format.

\paragraph{Minimal pairs.} From the test items in a fixed hash order, the AI coding agent that ran this experiment wrote \rPairsDir{} edits of
$O_{\text{new}}$ that turn a \SUP{} pair into \NC{} by breaking one contract condition (\rAPlanN{} of them turn a current
or completed state into a plan, wish or hypothetical; the rest move the scope) and \rPairsDir{} that turn an \NC{} pair
into \SUP{} by stating the missing fact; \rSkipA{} and \rSkipB{} items had no one-condition edit and were skipped. Both
members were sent fresh to Jev. S3 flips its decision with the edit on \rFlipSThree{} of \rPairs{} pairs and S2 on
\rFlipSTwo{} (flip in only one gate: S3 \rFlipOnlyThree, S2 \rFlipOnlyTwo; McNemar $p=\rFlipP$), and no flip by either
went against the edit. On the \SUP{}-to-\NC{} edits S2 retires \rAOrigSTwo{} originals and still \rAEditSTwo{} edited
members, S3 \rAOrigSThree{} and \rAEditSThree; on the \NC{}-to-\SUP{} edits S2 retires \rBOrigSTwo{} originals and
\rBEditSTwo{} edited members, S3 \rBOrigSThree{} and \rBEditSThree. By the pre-registered rules S3 reads the edited
feature (at least 70\% flips, $p<.01$) and S2 does not, and S3 flips on more pairs by the fixed margin (15 points), so on
these pairs the binary score follows the suggestion more than the three-way decision does. Fresh and stored decisions on
the original items agree on \rStableSTwo{} (S2) and \rStableSThree{} (S3) of \rPairs. Caveats: that agent wrote every edit
knowing the hypothesis, and no human, second author or checker reviewed them (the plan asked for two authors and a third checker); the edited members have no
blind label; and most edits change exactly the feature S3's prompt names (plan or intent), so the test shows that S3
applies its stated rule to these texts, not how either gate reads natural data.

\section{Output format alone, and a blind relabel that was not run (pre-registered after the main results)}\label{app:r9}
A further plan was committed to version control before any call; it fixed the questions below, the sample and the
decision rules, and the item outputs were committed before they were scored.

\paragraph{Design.} Two Jev setups differ only in the output they ask for. Both send S3's state (the claim, $O_{\text{old}}$,
$O_{\text{new}}$ and both times) and S3's instruction word for word. One asks S3's Choice over the three contract options.
The other asks a yes/no Noul whose yes is S3's \SUP{} definition and whose no is its \NC{} and \COMP{} definitions
together, each headed by its label name, and retires at $\tau{=}.5$. A second binary form asks a Choice over two options
with the same texts. S2, with its own prompt and without $O_{\text{old}}$, was re-called alongside as a reference whose
purpose does not match. No STALE scenario was left untouched: the \stUnused{} scenarios
outside our splits were used in the pilot work in which S2's prompt was written, and none has contract labels. The
comparison therefore runs on the \nTest{} test items, with every question fixed before any call; the test labels and the
test rates above were known to the authors when the plan was written. Each arm sent one fresh request per item, and the
yes/no and three-way arms sent a second one on the \mxRepN{} \SUP{} and \NC{} items. The primary rule reads the paired
difference in $P$ on the \nTestNC{} \NC{} items. If McNemar gives $p<.05$, the three options lower or raise $P$, as the
discordant pairs point; if $p\ge.05$ and the 95\% interval of the difference \citep[method~10]{newcombe1998paired} lies
within $\pm10$ points, they change it by less than 10 points; otherwise the comparison is unresolved.

\begin{table}[h]
\centering\small
\caption{\textbf{Output format alone} (test, $n{=}\nTest$; rates in \%, Wilson 95\% intervals). The first three rows share
S3's prompt, state and model and differ only in the requested output; $F_{\text{ws}}$ counts the \mxWsN{}
same-conversation controls.}
\label{tab:e9a}
\begin{tabular}{lccccc}
\toprule
Output & $D$ & $P$ & $F$ & $F_{\text{ws}}$ & AUROC \\
\midrule
three options (Choice) & 77.8 {\scriptsize [66.9, 85.8]} & 17.5 {\scriptsize [11.2, 26.3]} & 1.2 {\scriptsize [0.5, 3.1]} & 0/141 & .956 \\
yes/no (Noul, $\tau{=}.5$) & 79.2 {\scriptsize [68.4, 86.9]} & 18.6 {\scriptsize [12.1, 27.4]} & 1.2 {\scriptsize [0.5, 3.1]} & 0/141 & .957 \\
two options (Choice) & 75.0 {\scriptsize [63.9, 83.6]} & 15.5 {\scriptsize [9.6, 24.0]} & 1.2 {\scriptsize [0.5, 3.1]} & 0/141 & .963 \\
binary gate, own prompt (fresh) & 93.1 {\scriptsize [84.8, 97.0]} & 38.1 {\scriptsize [29.1, 48.1]} & 1.2 {\scriptsize [0.5, 3.1]} & 0/141 & .965 \\
\bottomrule
\end{tabular}

\end{table}

\paragraph{Result.} With the prompt and input fixed, a yes/no output retires \mxYNPreK/\nTestNC{} suggested changes and
three options \mxThreePreK/\nTestNC{} (McNemar \mxMcA:\mxMcB, $p=\mxMcP$; difference \mxDiff{} points, 95\% interval
[\mxDiffLo, \mxDiffHi]; \cref{tab:e9a}), so by the plan's rule the output format alone changes premature retirement by
less than 10 points. Detection is \mxYNDetK{} and \mxThreeDetK{} of \nTestSup{} (McNemar \mxDetA:\mxDetB, $p=\mxDetP$).
For this matched pair the three-way hypothesis fails as well, now because the three options do not lower $P$. Changing the
prompt and input does: S2, re-called, retires \mxSTwoPreK/\nTestNC{} against the yes/no output's \mxYNPreK{} (McNemar
\mxSTwoA:\mxSTwoB, $p<.001$), and the output accounts for \mxShare\% of the gap between S2 and the three options. At
matched detection ($\tau{=}\mxTau$, \mxTauDetK/\nTestSup{} for both) the yes/no output retires \mxTauPreK/\nTestNC{}
(McNemar \mxTauA:\mxTauB, $p=\mxTauP$), and the two-option Choice retires \mxYNTwoPreK/\nTestNC{} (\mxCA:\mxCB,
$p=\mxCP$). Sent a second time, the yes/no output repeats \mxRepYN{} and the three options \mxRepThree{} of \mxRepN{}
decisions, and the rule's reading is unchanged (\mxRepYNPreK{} and \mxRepThreePreK{} of \nTestNC). Fresh and stored
decisions agree on \mxStabSTwo{} (S2) and \mxStabThree{} (S3) of \mxStabN{} items. The \mxCalls{} requests, two of them a format check on one development item,
cost \$\mxUSD. Both matched outputs meet the safety bar on point estimates, with $P$ intervals that reach past .20
([\mxYNPreLo, \mxYNPreHi] and [\mxThreePreLo, \mxThreePreHi]). Caveats: these are test items the authors had scored
before, with one model on one benchmark, and the remaining wording choices (label names at the head of the binary
definitions) were fixed, not varied.

\paragraph{No independent blind relabelling.} No independent blind relabelling was done. The plan also fixed a relabel
by a person who is not an author, has not worked on the task and uses no AI tool; we prepared it but did not run it,
because no annotator meeting these conditions was available. Its \ebItems{} items are the \ebA{} \NC{} test items with LLM-majority gold,
the \ebB{} second-annotator items, and \ebCSup{} \SUP{} and \ebCComp{} \COMP{} test items with LLM-majority gold drawn at
random, with \ebRep{} items shown twice (\ebCards{} cards). The offline page holds no model output and no earlier label,
logs when each card is shown and submitted, and records the annotator's attestation. The plan fixes the agreement
statistics, the re-scored gates and a provenance check (timing, a browser-automation flag and a search of the authors'
coding-agent logs for the page) before any label exists. No such label exists, and no number in this paper uses one.
The \nReadj{} second-annotator labels therefore keep the provenance given in \cref{sec:data}: the second annotator is
now one of the authors, and whether they were one when they labelled is not established; the written condition that
this annotator not have worked on the task is not met; their retire decisions match the discarded AI pre-fill on all
\nReadj{} items (\retExtCodex/\nReadj); and every vault result is exploratory.

\end{document}